\documentclass[journal]{IEEEtran}

\usepackage{graphicx}
\usepackage{caption}
\usepackage{subfigure}
\usepackage{algpseudocode}
\usepackage{algorithmicx,algorithm}
\usepackage{bm}
\usepackage{booktabs}
\usepackage{cite}
\usepackage{stfloats}
\usepackage{amsmath}
\usepackage{makecell}
\usepackage{color}

\begin{document}

\title{Object Tracking by the Least Spatiotemporal Searches}

\author{Zhiyong~Yu,~\IEEEmembership{Member,~IEEE,} Lei~Han, Chao~Chen, Wenzhong~Guo, and~Zhiwen~Yu,~\IEEEmembership{Senior~Member,~IEEE}% <-this % stops a space
\thanks{Z. Yu and W. Guo are with College of Mathematics and Computer Sciences, Fuzhou University, Key Laboratory of Spatial Data Mining and Information Sharing, Ministry of Education, and Fujian Key Laboratory of Network Computing and Intelligent Information Processing, Fuzhou 350108, China. E-mail: {yuzhiyong, guowenzhong}@fzu.edu.cn.}% <-this % stops a space
\thanks{L. Han and Zhiwen Yu (corresponding author) is with School of Computer Science, Northwestern Polytechnical University, Xi’an 710072, China. E-mail: {hanlei, zhiwenyu}@nwpu.edu.cn.}
\thanks{C. Chen is with School of Computer Science, Chongqing University, Chongqing 400044, China. E-mail: cschaochen@cqu.edu.cn.}
\thanks{Manuscript received XX XX, XX; revised XX XX, XX.}}

% The paper headers
\markboth{Journal of \LaTeX\ Class Files,~Vol.~14, No.~8, August~2015}%
{Shell \MakeLowercase{\textit{et al.}}: Bare Demo of IEEEtran.cls for IEEE Journals}

% make the title area
\maketitle

% As a general rule, do not put math, special symbols or citations
% in the abstract or keywords.
\begin{abstract}
Tracking a suspicious car or a person in a city efficiently is crucial in urban safety management. But how can we complete the task with the minimal number of spatiotemporal searches when massive camera records are involved? To this end, this study proposes a strategy named intermediate searching at heuristic moments (IHMs). At each step, we determine which moment is the best one for the search based on a heuristic indicator. Then, at that moment, locations are searched one by one in descending order of predicted appearing probabilities until a search hit is obtained. We iterate this step until we derive the object’s current location. Five searching strategies are compared via experiments. Among these strategies, the IHMs strategy is validated as the most efficient. IHMs can save up to 1/3 of the total cost. This result provides evidence that “searching at intermediate moments can save cost.”
\end{abstract}

% Note that keywords are not normally used for peerreview papers.

\IEEEpeerreviewmaketitle

\section{Introduction}
\label{section1}

\IEEEPARstart{T}{racking} a car or a person in a city is crucial in urban safety management \cite{shin2014cosmic}. For instance, a suspicious car is witnessed at a location at a past time (witnessed moment, e.g., 14:00) but is out of sight shortly. Afterwards (current moment , e.g., 15:00), a police captures this clue and intends to monitor the car’s whereabouts to determine whether this car needs to be controlled immediately. However, the police cannot conveniently determine the car’s current location. The police may not be able to locate the car by GPS or a direct phone call. Thus, the police may decide to find the car through cameras that had been previously placed all over the city. 

The quantity of cameras can be up to several millions, and their records can last from a witnessed moment to the current moment. In the example given above, the challenge is “how can we find the car with a minimal number of spatiotemporal searches?” A spatiotemporal search can be described as the effort of the police to check whether the car appears at a specified location at a specified moment. Furthermore, all camera records are assumed to have been indexed according to their locations and moments. For simplicity, the unit cost of a spatiotemporal search can be considered a constant with regard to either human or artificial intelligence. The significance of solving this problem lies in the potential to save on total costs (total costs = amount of searches × unit cost) to track an object (i.e., to find where the object is at present).

\begin{figure*}[ht]
	\centering
	\begin{minipage}[t]{16cm}
		\setlength{\abovecaptionskip}{0.2cm}   %调整图片标题与图距离Fig. \ref{figure1}
		\setlength{\belowcaptionskip}{-0.25cm}   %调整图片标题与下文距离
		%		\subfigcapskip=-0.3cm %设置子图与子标题之间的距离
		\centering 
		\includegraphics[width=12cm]{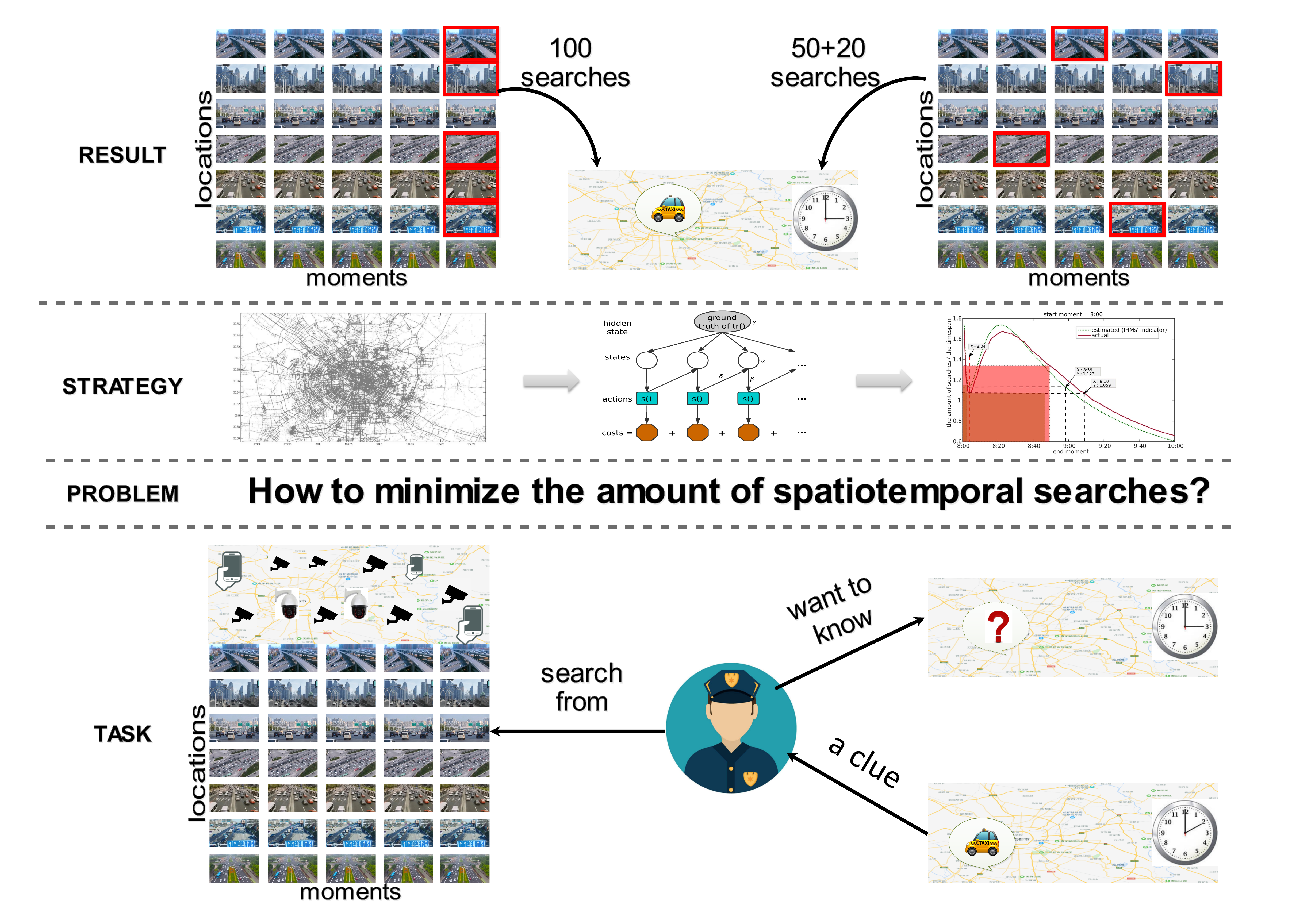}
		\caption{Framework of spatiotemporal searches}
		\label{figure1}
	\end{minipage}
\end{figure*}

A consensus is that we should first search where the object is most likely to be, which is the ability of mobility prediction. Among many models, the Markov model is a good choice for predicting mobility \cite{pecher2016data}. This model can be trained from historical trajectories, and it can provide probabilities of an object appearing at each location in the future. Besides the straightforward strategy in which all searches are executed at the last time (i.e., the current moment), another strategy worthy of exploring is performing searches at intermediate moments to gather additional locations where the object has passed through, to better predict where the object will be; thus, at the last time, we only need to perform fewer searches to find the object. As for the abovementioned example, the police would originally need 100 searches at the last time to find the suspicious car, but he or she might only need 20 more searches at the last time if 50 searches had already been spent at intermediate moments (Fig. \ref{figure1}).

In this research, we propose a strategy for object tracking by using the least spatiotemporal searches and compare it with other strategies. Then, we analyze whether the cost spent in the earlier searching can be compensated by the costs saved in the latter searching. The contributions of this study can be summarized as follows:

(1) In Section \ref{section2}, we define a novel problem of minimizing the number of spatiotemporal searches that can track an object after a specific time since its last appearance.

(2) In Section \ref{section3}, we propose a strategy with intermediate searching to solve the problem and determine when and where to search according to a heuristic indicator, which is estimated from historical trajectories. The heuristic indicator can resolve the conflict arising from spending the \emph{least} searches to identify the object’s \emph{latest} location. Other baseline strategies are also designed.

(3) In Section \ref{section4}, we evaluate our strategy by using with real-world cars’ trajectories, and validate that the idea of saving costs through intermediate searching is feasible. We find that the proposed strategy is better than the other strategies.

The related work is presented in Section \ref{section5}, and the conclusions are provided in Section \ref{section6}.

%\hfill mds
% 
%\hfill August 26, 2015

\section{Related Work}
\label{section5}

Corresponding to two innovations of this work: the novel problem, and the intermediate searching strategy, in this section, we elaborate why they are challenging by comparing with their existing counterparts respectively. Besides, we encapsulate related algorithms about time-specific mobility prediction.

\subsection{Problem and Approach of Searching at the Last Time}

Tracking an object that is out of sight for a short time is similar to searching a missing object. In the operational research community, Koopman \cite{1957The}, Stone \cite{Rota1976Theory}, and other researchers have developed the theory of optimal search from the perspective of military, rescue, and law enforcement application. For example, the Bayesian search theory has been used to locate the remains of Malaysia Airlines Flight 370 \cite{MH370}. The Bayesian search theory first formulates reasonable hypotheses as to where the object is located and constructs a probability density function accordingly. Then, it starts to search from the locations with the highest probability to the lower probability and revises the probabilities continuously during the search. ALT (All Searching at the Last Time) can be deemed as an existing approach that applies the Bayesian search theory. New challenges arise for the novel problem defined in this study. First, the traditional theory mostly handles a one-off planning problem that has a low rate of knowledge updating. By contrast, in our problem, the output of each search can immediately affect the decision making of the next search. Second, the traditional theory merely involves spatial searches at the current moment. In our problem, we can adopt spatiotemporal searches that can search at intermediate moments with the help of digital records, e.g., videos, photos, or check-ins \cite{2013Learning}. Third, the traditional theory lacks the perspective from real-world data (e.g., prefers theoretical analysis and simulation). By contrast, in our work, real-world data are used to train the mobility prediction model, estimate the costs, and evaluate the different strategies.

\subsection{Informative Searching at Intermediate Moments}

In this work, we focus on different searching strategies under the same mobility prediction algorithm, as well as provide an evidence that searching at intermediate moments can save cost. By contrast, the existing studies focus on different mobility prediction algorithms under the same searching strategy, i.e., ALT.

The idea of gathering more past locations to improve the prediction of future locations seems to be similar with active learning \cite{nguyen2013wi, 2018RADAR, 2014Active}. However, in active learning, the features of all samples and the labels of a few samples are known, then users decide which among the unlabeled samples are to be labeled next. For the unlabeled sample in this study, we lack its features and need to decide which feature is to be gathered next. Furthermore, in this study, we have to carry out much more work other than the traditional feature selection because, on the one hand, we assume that we cannot query directly an object’s past locations; thus, past locations cannot be candidate features to be selected. On the other hand, if we regard “whether an object was at a location at a moment” as a feature, then the features will be highly multi-collinear and unbalanced.

Since the strategy of spatiotemporal searches can be represented by a Markov decision process (MDP), and MDP is usually considered as the problem model of reinforcement learning, thus reinforcement learning would be a solution for spatiotemporal searches. This is a promising future work indeed. However, as a simple reinforcement learning model, multi-armed bandit is not suitable for our problem. In a multi-armed bandit \cite{vermorel2005multi}, each arm has some unknown probability of dispensing a unit of reward, and the only decision we can make is which arm to pull next, in order to maximize the total rewards with certain times of pulling. While in the problem of spatiotemporal searches, when and where to search next is a decision, but “when and where”, i.e., a spatiotemporal unit, does not have unknown probability after one search, because the feedback is the object is found or not. There is no need for further exploration or exploitation in that unit.

\subsection{Time-Specific Mobility Prediction}

Few researchers have dedicated their focus on time-specific mobility prediction. Nevertheless, destination prediction \cite{xu2016destpre, 2017Moving, 2010A, 2006Predestination}, next-place prediction \cite{10.1145/2783258.2783350, 2014Limits, 2020A}, and trajectory compression \cite{2020TrajCompressor} can be extended by integrating time prediction to handle time-specific mobility prediction. In our previous work \cite{Yu2016Multi}, we proposed a Markov-based time-specific mobility prediction algorithm that could outperform the next-place based algorithm. The proposed technique considered multiple factors, including personal habit, weekday similarity, and collective behavior \cite{2017Toward}. In the present study, we exploit several variants of this algorithm take into account the past moment or past locations of the object.

\section{Problem Definition}
\label{section2}

%\newcommand{\tabincell}[2]{\begin{tabular}{@{}#1@{}}#2\end{tabular}} 
%表格
\begin{table}[!htb]
	\centering
	\footnotesize
	\caption{Notations}
	\label{tab1}
	\setlength{\tabcolsep}{1.5mm}{
		\begin{tabular}{p{50pt}<{\centering}|p{170pt}}
			\toprule
			Symbol & \makebox[6cm][c]{Description} \\
			\midrule
			${d_j}$ & a day,  the ${d_j} \in $ set of days $D$; ${d_x}$ is the testing day \\
			${t_k}$ & a moment in a day, ${d_j} = \left\langle {{t_1},{t_2},...,{t_{|{d_j}|}}} \right\rangle $; particularly ${t_p}$ is the start moment, ${t_x}$ is the end moment \\
			${c_x}$ & an object, ${c_x} \in $ the set of objects $C$ \\
			${l_i}$ & a location, ${l_i} \in $ the set of locations $L$; ${l_{tk}}$ means the time-specific location at ${t_k}$ \\
			$({l_i},{t_k})$ & a spatiotemporal unit \\
			$tr({c_x},{d_j})$ & a trajectory, $tr({c_x},{d_j}) = \left\langle {({l_{{t_1}}},{t_1}),({l_{{t_2}}},{t_2}),...} \right\rangle |{c_x},{d_j}$ \\
			$TR$ & the set of trajectories, $TR = \{ tr({c_x},{d_j})\} $ \\
			$TR'$ & the set of history trajectories, i.e., before the testing day \\
			$s({c_x},{l_i},{t_k})$ & the result of a spatiotemporal search, 1 for success, 0 for failure \\
			$p({l_i},{l_j})|\Delta t$ & the probability of an object that moves from one location to another after a period of time $\Delta t:{t_p} \to {t_x}$ \\
			$TPM_{\Delta t}^{|L| \times |L|}$ & a transition probability matrix ($TPM$); an element is $p({l_i},{l_j})|\Delta t$, $p({l_i},{l_j})$ for brevity \\
			$\mathop p\limits^ \to  ({c_x},{t_k})$ & the result of the mobility prediction algorithm, e.g., a row of a $TPM$, which moves from $l_i$ \\
			${p({l_j})}$ & the probability of the object $c_x$ will locate at $l_j$ at the moment $t_k$, an element of $\mathop p\limits^ \to  ({c_x},{t_k})$ \\
			$\mathop {p'}\limits^ \to  ({c_x},{t_k})$ & the result of sorting the elements of $\mathop p\limits^ \to  ({c_x},{t_k})$ in descending order \\
			$p(l{'_j})$ & an element of $\mathop {p'}\limits^ \to  ({c_x},{t_k})$, $p(l{'_1}) \ge p(l{'_2}) \ge ... \ge p(l{'_{|L|}})$ \\
			${\mathord{\buildrel{\lower3pt\hbox{$\scriptscriptstyle\frown$}} 
					\over m} }$, ${\mathord{\buildrel{\lower3pt\hbox{$\scriptscriptstyle\frown$}} 
					\over n} }$, $en( \cdot )$ & the estimated amount of searches needed \\
			$m$, $n$ & the actual amount of searches \\
			ALT & All Searching at the Last Time \\
			IPM & Intermediate Searching at a Parametric Moment \\
			IEM & Intermediate Searching at an Estimated Moment \\
			IHMs & Intermediate Searching at Heuristic Moments \\
			IHUs & Intermediate Searching at Heuristic Spatiotemporal Units \\
			\bottomrule
	\end{tabular}}
\end{table}

The notations listed in Table \ref{tab1} are used to formally describe the problem of “tracking an object by the least spatiotemporal searches.” Assume that the city can be discretized into non-overlapping locations and time can be discretized into non-overlapping moments. Let $C = \{ {c_1},{c_2},...,{c_{|C|}}\} $ be objects (e.g., cars or persons); $L = \{ {l_1},{l_2},...,{l_{|L|}}\} $ be locations of a real-world area (e.g., 1 km × 1 km grids); $D = \{ {d_1},{d_2},...,{d_{|D|}}\} $ be days, where ${d_j} = \left\langle {{t_1},{t_2},...,{t_{|{d_j}|}}} \right\rangle $ and ${t_k}$ is each moment (e.g., a minute); and $TR = \{ tr({c_i},{d_j})\} $ be trajectories, where $tr({c_i},{d_j}) = \left\langle {({l_{{t_1}}},{t_1}),({l_{{t_2}}},{t_2}),...} \right\rangle |{c_i},{d_j}$. In this manner, the \emph{trajectory} can be defined as the sequence of spatiotemporal units representing time-specific locations of a day of an object. Here, an object is regarded to be only at one location at a moment. If an object is extremely fast and passes through several locations in a minute, then we only need to preserve the first location of that minute.

\textbf{Definition 1} (spatiotemporal search): From the camera records of day ${d_j}$, a spatiotemporal search takes an object ${c_x}$, a location ${l_i}$, and a moment ${t_k}$ as the inputs and outputs a Boolean value, which is 1 if the object locates ${l_i}$ at ${t_k}$ or 0 otherwise.

\begin{small}
	\begin{equation}
	\begin{split}
	s({c_x},{l_i},{t_k}) = \left\{ {\begin{array}{*{20}{c}}
		{1,}\\
		{0,}
		\end{array}} \right.\begin{array}{*{20}{c}}
	{({l_i},{t_k}) \in tr({c_x},{d_j})}\\
	{else}
	\end{array}
	\end{split}
	\label{eq1}
	\end{equation}
\end{small}

Remind that camera records are assumed to have been indexed according to their locations and moments, but not indexed by contents, e.g., license number, because content recognition is only performed on the records of a specified location at a specified moment when the “searching” is conducted within a certain indexed content.

The unit cost of a search depends on the granularity of the spatiotemporal unit, the volume and quality of the camera records, and the technique of content recognition. As the target object needs to be located in a short time (e.g., within the current 1 minute), the unit cost is mainly measured using other resources, such as the number of agents working in parallel, instead of time length.

\textbf{Problem 1} (tracking an object by the least spatiotemporal searches): The following are known: an object ${c_x}$, a day ${d_x}$, a past moment ${t_p}$ of that day, ${c_x}$’s location ${l_{{t_p}}}$ at that moment, a current moment ${t_x}$, history trajectories $TR'$ (in which arbitrary ${d_j} < {d_x}$), and the camera records that allow the spatiotemporal search $s({c_x},{l_i},{t_k})$ to be performed in ${d_x}$ before/at ${t_x}$. The knowledge as to when and where to execute a spatiotemporal search can be determined on the basis of the $TR'$ and the outputs of the previous searches. The problem is how to utilize the minimal amount of searches to find the object at the current moment ${t_x}$. 

\begin{small}
	\begin{equation}
	\begin{split}
	{\mathop{\rm \textbf{give}}\nolimits}\  {c_x},{d_x},{t_x},({l_{{t_p}}},{t_p}),TR'\ {\mathop{\rm before}\nolimits}\  {d_x},s({c_x},{l_i},{t_k})\ {\mathop{\rm in}\nolimits}\  {d_x}\\
	\quad{\mathop{\rm \textbf{return}}\nolimits}\  {l_x},s.t.\mathop {\min }\limits_{s({c_x},{l_x},{t_x}) = 1} |\left\langle {s(),s(),...,s({c_x},{l_x},{t_x})} \right\rangle |
	\end{split}
	\label{eq2}
	\end{equation}
\end{small}

\section{Strategies of Spatiotemporal Searches}
\label{section3}

\begin{figure}[ht]
	\centering
	\begin{minipage}[t]{8cm}
		\setlength{\abovecaptionskip}{0.2cm}   %调整图片标题与图距离Fig. \ref{figure1}
		\setlength{\belowcaptionskip}{-0.25cm}   %调整图片标题与下文距离
		%		\subfigcapskip=-0.3cm %设置子图与子标题之间的距离
		\centering 
		\includegraphics[width=6cm]{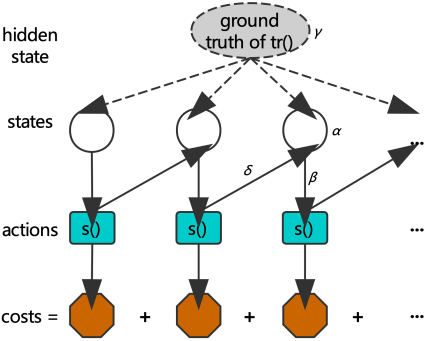}
		\caption{Markov decision process of spatiotemporal searches}
		\label{figure2}
	\end{minipage}
\end{figure}

The interaction between the problem and the solution (i.e., the strategy of spatiotemporal searches) can be represented by a Markov decision process, as shown in Fig. \ref{figure2}. At each step, the process is in a state denoting how much we know about the object’s trajectory, and the strategy chooses an action denoting when and where to search on the basis of the current state. The process responds at the next step by moving into a new state, and a corresponding cost is incurred. From this analysis, we can presume that the total costs will be determined by the searching difficulty $\alpha $ (influenced by the predictability of the object’s trajectory $\gamma $ and the effectiveness of the mobility prediction algorithm $\delta $) and the efficiency of the searching strategy $\beta $.

\subsection{Mobility Prediction Algorithm}
We introduce why and how the first-order Markov model is applied to predict an object’s future time-specific location. Nonetheless, the mobility prediction algorithm is not the focus and contribution of this study. Other algorithms (e.g., deep learning-based algorithms \cite{song2016deeptransport, wu2017modeling}) are not adopted in this work because, first, a spatiotemporal search involves multiple rounds of “predicting and searching,” which may lead to the costly training of a deep learning model. Second, the known locations are extremely sparse, especially at the beginning of an algorithm, which can hinder the deep learning model from fully realizing its power. The Markov model can attain good performance for the above mentioned situation \cite{pecher2016data}. More effective mobility prediction algorithms will be explored in the future.

Let a transition probability matrix ($TPM$) consist of the probabilities of an object moving from one location to another after a period of time $\Delta t$, i.e.,

\begin{small}
	\begin{equation}
	\begin{split}
	TPM_{\Delta t}^{|L| \times |L|} = \{ p({l_i},{l_j})|\Delta t\}, 
	\end{split}
	\label{eq3}
	\end{equation}
\end{small}

\begin{small}
	\begin{equation}
	\begin{split}
	p({l_i},{l_j})|\Delta t = \frac{{\# tr(){\mathop{\rm with}\nolimits} ({l_i},{t_p})\& ({l_j},{t_p} + \Delta t)}}{{\# tr(){\mathop{\rm with}\nolimits} ({l_i},{t_p})}},\\
	{t_p} = {t_1},{t_2},...,tr() \in TR'
	\end{split}
	\label{eq4}
	\end{equation}
\end{small}

Then, $TPM_{\Delta t}^{|L| \times |L|}$ can be used to predict any object’s location of the time $\Delta t = {t_x} - {t_p}$ later. The $TPM$ describes that how possible an object transits from one location to another location. It is not only one value but a matrix, which means each location pair is calculated separately. The values are the rates of statistical frequency from history trajectories. Since history trajectories are influenced by traffic conditions of those blocks of those time, so we say traffic conditions (not real-time) are already implicitly considered in Markov model.

For instance, a car was present at a ${l_{45}}$ at 8:00 am today, and it is now 8:30 am. Its current location will most probably be ${l_{46}}$, knowing that $p({l_{45}},{l_{46}})|30\min  = 0.51$, with a second probability at ${l_{47}}$, knowing that $p({l_{45}},{l_{47}})|30\min  = 0.32$, and so on, until all locations are enumerated exhaustively.

The calculation process of Equation (\ref{eq4}) and those of other equations are performed using toy examples with five trajectories, four locations, and three moments. The dataset is shown in Table \ref{tab2}. The equations of the toy examples are not numbered for discrimination.

\begin{table}[!htb]
	\centering
	\footnotesize
	\caption{Dataset of the Toy Examples}
	\label{tab2}
	\setlength{\tabcolsep}{1.5mm}{
		\begin{tabular}{p{40pt}<{\centering}|p{20pt}<{\centering} p{20pt}<{\centering} p{20pt}<{\centering}}
			\toprule
			 & ${t_1}$ & ${t_2}$ & ${t_3}$ \\
			\midrule
			$tr({c_1},{d_1})$ & ${l_1}$ & ${l_2}$ & ${l_3}$ \\
			$tr({c_1},{d_2})$ & ${l_2}$ & ${l_2}$ & ${l_3}$ \\
			$tr({c_1},{d_3})$ & ${l_2}$ & ${l_1}$ & ${l_1}$ \\
			$tr({c_2},{d_1})$ & ${l_3}$ & ${l_4}$ & ${l_4}$ \\
			$tr({c_2},{d_2})$ & ${l_2}$ & ${l_4}$ & ${l_3}$ \\
			\bottomrule
	\end{tabular}}
\end{table}

Let ${t_p} = {t_1}$, with $\Delta t$ representing a value from ${t_1}$ to ${t_3}$, then we calculate $4 \times 4 = 16$ probabilities in $TPM$. Here, we only demonstrate the calculation of $p({l_2},{l_3})$ according to Equation (\ref{eq4}).

\[
\begin{split}
p({l_2},{l_3})|\Delta t = \frac{{\# tr()\ {\mathop{\rm with}\nolimits}\  ({l_2},{t_1})\& ({l_3},{t_3})}}{{\# tr()\ {\mathop{\rm with}\nolimits}\  ({l_2},{t_1})}}\qquad \qquad \quad\ \ \,\\
= \frac{{|\{ tr({c_1},{d_2}),tr({c_2},{d_2})\} |}}{{|\{ tr({c_1},{d_2}),\{ tr({c_1},{d_3}),\{ tr({c_2},{d_2})\} |}} = \frac{2}{3}
\end{split}
\]

The meaning of “$\# $” is “the number of.” Thus, this example pertains to the obtainment of probabilities in which objects would transit from location ${l_2}$ to location ${l_3}$ after two moments. We statistically count the frequency of historical trajectories that occurred to meet the two conditions (${l_2}$ at ${t_1}$ AND ${l_3}$ at ${t_3}$ as the numerator, and ${l_2}$ at ${t_1}$ as the denominator), divide them, and so on, until each $p({l_i},{l_j})$ is calculated. Then, we calculate the $TPM$ as follows:

\[
\begin{split}
TP{M_{{t_1} \to {t_3}}} = \left( {\begin{array}{*{20}{c}}
	0&0&1&0\\
	{1/3}&0&{\textbf{2/3}}&0\\
	0&0&0&1\\
	{1/4}&{1/4}&{1/4}&{1/4}
	\end{array}} \right)
\end{split}
\]

If the denominator is 0, then we reset the probability to 1/4, with the simple assumption that it will equally distribute at all four locations, as shown in the last row of the above $TPM$. For other cases, smoothing is not used because the default “+1 smoothing” may change the probabilities drastically and unevenly in our scenario, or it needs non-default hyperparameters which are hard to tune. This could also be a future work (but not the focus of the current one) to improve the mobility prediction algorithm.

In general, knowing more locations through which the object has passed will bring higher prediction accuracy. However, the results will depend on sufficient data. If the matched history trajectories for training are insufficient, then the high-order Markov model will be inaccurate because of low signal-to-noise ratio (SNR) or overfitting. Thus, we only take the second-order Markov model into consideration. Its $TPM$ (transition from ${l_i}$ to ${l_j}$ through ${l_h}$) is calculated as follows:

\begin{small}
	\begin{equation}
	\begin{split}
	TPM_{\Delta t,{l_h}}^{|L| \times |L|} = \{ p({l_i},{l_j})|\Delta t,{l_h}\} 
	\end{split}
	\label{eq5}
	\end{equation}
\end{small}

\begin{small}
	\begin{equation}
	\begin{split}
	p({l_i},{l_j})|\Delta t,{l_h} = \frac{{\# tr()\ {\mathop{\rm with}\nolimits}\  ({l_i},{t_p})\& ({l_h},{t_k})\& ({l_j},{t_p} + \Delta t)}}{{\# tr()\ {\mathop{\rm with}\nolimits}\  ({l_i},{t_p})\& ({l_h},{t_k})}},\\
	{t_k} \in [{t_p},{t_p} + \Delta t],tr() \in TR'
	\end{split} 
	\label{eq6}
	\end{equation}
\end{small}

Similar to the toy example in Equation (\ref{eq4}), we can calculate Equation (\ref{eq6}) when we know if the object was at location ${l_4}$ at ${t_2}$. The transition probability from ${l_2}$ to ${l_3}$ is given by

\[
\begin{split}
p({l_2},{l_3})|\Delta t,{l_4} = \frac{{\# tr()\ {\mathop{\rm with}\nolimits}\  ({l_2},{t_1})\& ({l_4},{t_2})\& ({l_3},{t_3})}}{{\# tr()\ {\mathop{\rm with}\nolimits}\  ({l_2},{t_1})\& ({l_4},{t_2})}}\\
\qquad \qquad \qquad \,= \frac{{|\{ tr({c_2},{d_2})\} |}}{{|\{ tr({c_2},{d_2})\} |}} = 1
\end{split}
\]

After all $p({l_i},{l_j})$ are calculated, we compute the $TPM$ as follows:

\[
\begin{split}
TP{M_{{t_1} \to {t_3},{(_{{l_4},{l_2}}})}} = \left( {\begin{array}{*{20}{c}}
	{1/4}&{1/4}&{1/4}&{1/4}\\
	0&0&\textbf{1}&0\\
	0&0&0&1\\
	{1/4}&{1/4}&{1/4}&{1/4}
	\end{array}} \right)
\end{split}
\]

\subsection{Expected or Estimated Amount of Searches}

Suppose the result of the mobility prediction algorithm is in the form of a vector (e.g., a row of $TPM$), then $\mathop p\limits^ \to  ({c_x},{t_x}) = \left\langle {p({l_1}),p({l_2}),...,p({l_{|L|}})} \right\rangle $, in which each element denotes the probability of object ${c_x}$ to be located at ${l_j}$ at moment ${t_k}$. We can calculate the expected amount of searches ($en$) needed to find ${c_x}$ at ${t_k}$ in $TR'$ in this manner: sort the elements of $\mathop {p'}\limits^ \to  ({c_x},{t_x}) = \left\langle {p(l{'_1}),p(l{'_2}),...,p(l{'_{|L|}})} \right\rangle $, where $p(l{'_1}) \ge p(l{'_2}) \ge ... \ge p(l{'_{|L|}})$, then we compute Equation (\ref{eq7}). The reason for this approach is that the location with the largest probability only needs one search and hit, but the location with the second largest probability will be searched and hit after a failed search, and so on. According to the law of large numbers (under the condition that the training and testing data are abundant and identically distributed), this scheme can also represent the estimated amount of searches needed to find ${c_x}$ at ${t_k}$ in ${d_x}$ (note that the ground truth of ${d_x}$’s trajectory is only for performance evaluation and remains unknown in the process of real-world spatiotemporal searching).

\begin{small}
	\begin{equation}
	\begin{split}	
	en(\mathop {p'}\limits^ \to  ({c_x},{t_k})) = \sum\nolimits_{j = 1}^{|L|} {j \times \mathop {p'}\limits^ \to  ({c_x},{t_k})[j]}  = \sum\nolimits_{j = 1}^{|L|} {j \times p(l{'_j})} 
	\end{split}
	\label{eq7}
	\end{equation}
\end{small}

As a continuation of the toy examples, we then search an object ${c_x}$ at $t_3$ knowing that its location is $l_2$ at $t_1$. As the second row of $TP{M_{{t_1} \to {t_3}}}$ is given by $\mathop p\limits^ \to  ({c_x},{t_3}) = \left\langle {p({l_1}) = 1/3,p({l_2}) = 0,p({l_3}) = 2/3,p({l_4}) = 0} \right\rangle $, it can be sorted into $\mathop {p'}\limits^ \to  ({c_x},{t_3}) = \left\langle {p({l_3}) = 2/3,p({l_1}) = 1/3,p({l_2}) = 0,p({l_4}) = 0} \right\rangle $. According to Equation (\ref{eq7}),

\[
\begin{split}
en(\mathop {p'}\limits^ \to  ({c_x},{t_3})) = 1 \times (2/3) + 2 \times (1/3) + 3 \times 0 + 4 \times 0{\rm{ = }}4/3
\end{split}
\]

We expect to find the object with $4/3$ searches because we had initially searched $l_3$, which has $2/3$ of probability to find it and costs 1 search; otherwise, search $l_1$, which has $1/3$ of probability to succeed and costs 2 cumulative searches, then $l_2$ and $l_4$.

\subsection{Five Searching Strategies}

The strategy of spatiotemporal searching aims to determine the location and the moment of each search in $\left\langle {s(),s(),...,s({c_x},{l_x},{t_x})} \right\rangle $. According to Definition 1 (spatiotemporal search), a search ${s({c_x},{l_i},{t_k})}$ has three parameters: object, location, and moment. If $c_y$ and $c_x$ are correlated, the traces of object $c_y$ may help to track object $c_x$. Then strictly speaking, knowing which object to search is another parameter that needs to be determined. However, in this study, we reasonably assume that target object $c_x$’s information can help most and the other objects’ information can be ignored. We introduce various strategies, including several baselines and a proposed one, which are detailed as follows.

\subsubsection{All Searching at the Last Time (ALT)}

A straightforward strategy is to execute all searches at the last time. Knowing the $({l_{{t_p}}},{t_p})$ of object $c_x$ for day $d_x$’s trajectory $tr({c_x},{d_x})$, the mobility prediction algorithm (e.g., first-order Markov model) yields descending-sorted probabilities $\mathop {p'}\limits^ \to  ({c_x},{t_x})$. This strategy called all searching at the last time (ALT) executes searches according to the sequence $\left\langle {s({c_x},l{'_1},{t_x}),s({c_x},l{'_2},{t_x}),...} \right\rangle $ until $s({c_x},l{'_n},{t_x}) = 1$, and $n$ is the actual amount of searches that has found $c_x$ at $t_x$ in $d_x$. In addition, the model can be trained offline. The time complexity of the offline training is $O(n)$; similarly, the time complexity of the online testing is $O(n)$.

\subsubsection{Intermediate Searching at a Parametric Moment (IPM)}

The steps of the IPM strategy are as follows:

(1) A parametric moment is specified and denoted as ${t_k} \in ({t_p},{t_x}]$.

(2) By knowing the $({l_{{t_p}}},{t_p})$ of $tr({c_x},{d_x})$, the mobility prediction algorithm (e.g., first-order Markov model) yields descending-sorted probabilities $\mathop {p'}\limits^ \to  ({c_x},{t_k})$.

(3) Searches at $t_k$ are executed from locations in the order of higher to lower probability until the object is found, i.e., $s({c_x},l{'_m},{t_k}) = 1$. (Steps 2 and 3 adhere to the ALT strategy.)

(4) By knowing $({l_{{t_p}}},{t_p})$ and $(l{'_m},{t_k})$ of $tr({c_x},{d_x})$, the mobility prediction algorithm (e.g., second-order Markov model or first-order Markov model with updated location) yields descending-sorted probabilities $\mathop {p'}\limits^ \to  ({c_x},{t_x})$.

(5) Finally, searches at $t_x$ are executed until the object is found, i.e., $s({c_x},l{'_n},{t_x}) = 1$, and $m + n$ is the actual amount of searches that has found $c_x$ at $t_x$ in $d_x$.

The model can be trained offline. The time complexity of the offline training is $O(n)$; similarly, the time complexity of the online testing is $O(n)$.

\subsubsection{Intermediate Searching at an Estimated Moment (IEM)}

The intermediate searching at an estimated moment (IEM) strategy is the same as the IPM strategy except for step 1. This strategy estimates the optimal moment that can minimize the total amount of searches. The optimal moment ${t_{opt}}$ is found by traversing all moments from $t_p$ to $t_x$. For each moment ${t_k} \in ({t_p},{t_x}]$, a simulated IPM strategy is rehearsed. As the actual amount of searches in the simulated IPM strategy remains unknown, we take an estimated amount of searches, i.e., ${\mathord{\buildrel{\lower3pt\hbox{$\scriptscriptstyle\frown$}} 
		\over m} }$ and ${\mathord{\buildrel{\lower3pt\hbox{$\scriptscriptstyle\frown$}} 
		\over n} }$ may be computed with Equation (\ref{eq7}) if possible, therefore avoiding the costs of searching massive camera records in this step. The object’s location at $t_k$ remains unknown, but we know that $\mathop p\limits^ \to  ({c_x},{t_k}) = \left\langle {p({l_1}),p({l_2}),...,p({l_{|L|}})} \right\rangle $. Thus, ${\mathord{\buildrel{\lower3pt\hbox{$\scriptscriptstyle\frown$}} 
		\over n} }$ is calculated as Equation (\ref{eq10}).
	
\begin{small}
	\begin{equation}
	\begin{split}
	{t_{opt}} = \mathop {\arg \min }\limits_{{t_k} \in ({t_p},{t_x}]} \mathop {(\mathord{\buildrel{\lower3pt\hbox{$\scriptscriptstyle\frown$}} 
			\over m}  + \mathord{\buildrel{\lower3pt\hbox{$\scriptscriptstyle\frown$}} 
			\over n} |{\mathop{\rm IPM}\nolimits} ({t_k}))}\limits_{}  
	\end{split}
	\label{eq8}
	\end{equation}
\end{small}

\begin{small}
	\begin{equation}
	\begin{split}
	\mathord{\buildrel{\lower3pt\hbox{$\scriptscriptstyle\frown$}} 
		\over m}  = en(\mathop {p'}\limits^ \to  ({c_x},{t_k})|tr({c_x},{d_x})\ {\mathop{\rm with}\nolimits}\  ({l_{{t_p}}},{t_p}))
	\end{split}
	\label{eq9}
	\end{equation}
\end{small}

\begin{small}
	\begin{equation}
	\begin{split}
	\mathord{\buildrel{\lower3pt\hbox{$\scriptscriptstyle\frown$}} 
		\over n}  = \sum\limits_{j = 1}^{|L|} {\left[ {p({l_j}) \times en\left( {\begin{array}{*{20}{c}}
				{\mathop {p'}\limits^ \to  ({c_x},{t_x})|}\\
				{tr({c_x},{d_x})\ {\mathop{\rm with}\nolimits}\  ({l_{{t_p}}},{t_p})({l_j},{t_k})}
				\end{array}} \right)} \right]}  
	\end{split}
	\label{eq10}
	\end{equation}
\end{small}

The IPM strategy is truly performed using ${t_{opt}}$ as the parametric moment.

As a continuation of the toy examples, when we want to estimate the amount of searches, and assuming that we know an extra location between the start moment and the end moment but the extra location is unknown during the estimation, we need to rewrite Equation (\ref{eq7}) to Equation (\ref{eq10}), in which ${p({l_j})}$ is the probability of the extra location. Then, we compute the weighted sum. For example, we begin to search an object cx at t3 knowing that its location is l2 at t1. To calculate Equation (\ref{eq10}), we first prepare $TP{M_{{t_1} \to {t_2}}}$, $TP{M_{{t_1} \to {t_3}({l_1},{t_2})}}$, $TP{M_{{t_1} \to {t_3}({l_2},{t_2})}}$, $TP{M_{{t_1} \to {t_3}({l_3},{t_2})}}$ and $TP{M_{{t_1} \to {t_3}({l_4},{t_2})}}$, although we only need the second row of these TPMs.

$\begin{array}{l}
\mathop p\limits^ \to  ({c_x},{t_2}) = \\
\begin{array}{*{20}{c}}
{}&{}
\end{array}\left\langle {p({l_1}) = 1/{\rm{3}},p({l_2}) = 1/3,p({l_3}) = 0,p({l_4}) = 1/3} \right\rangle 
\end{array}$

$\begin{array}{l}
\mathop p\limits^ \to  ({c_x},{t_3})|({l_1},{t_2}) = \\
\begin{array}{*{20}{c}}
{}&{}
\end{array}\left\langle {p({l_1}) = 1,p({l_2}) = 0,p({l_3}) = 0,p({l_4}) = 0} \right\rangle 
\end{array}$

$en( \cdot ) = 1$ by Equation (\ref{eq7})

$\begin{array}{l}
\mathop p\limits^ \to  ({c_x},{t_3})|({l_2},{t_2}) = \\
\begin{array}{*{20}{c}}
{}&{}
\end{array}\left\langle {p({l_1}) = 0,p({l_2}) = 0,p({l_3}) = 1,p({l_4}) = 0} \right\rangle 
\end{array}$

$en( \cdot ) = 1$ by Equation (\ref{eq7})

$\begin{array}{l}
\mathop p\limits^ \to  ({c_x},{t_3})|({l_3},{t_2}) = \\
\begin{array}{*{20}{c}}
{}&{}
\end{array}\left\langle {p({l_1}) = 1/4,p({l_2}) = 1/4,p({l_3}) = 1/4,p({l_4}) = 1/4} \right\rangle 
\end{array}$

$en( \cdot ) = 5/2$ by Equation (\ref{eq7})

$\begin{array}{l}
\mathop p\limits^ \to  ({c_x},{t_3})|({l_4},{t_2}) = \\
\begin{array}{*{20}{c}}
{}&{}
\end{array}\left\langle {p({l_1}) = 0,p({l_2}) = 0,p({l_3}) = 1,p({l_4}) = 0} \right\rangle 
\end{array}$

$en( \cdot ) = 1$ by Equation (\ref{eq7})

According to Equation (\ref{eq10}), we can obtain $\mathord{\buildrel{\lower3pt\hbox{$\scriptscriptstyle\frown$}} 
	\over n}  = 1/3 \times 1 + 1/3 \times 1 + 0 \times 5/2 + 1/3 \times 1 = 1$. Thus, if we know the object’s location at $t_2$, then we only need one more search to likely find it at $t_3$. In addition, the model can be trained offline. The time complexity of the offline training is $O({n^2})$, while the time complexity of the online testing is $O(n)$.

\subsubsection{Intermediate Searching at Heuristic Moments (IHMs)}

The intermediate searching at heuristic moments (IHMs) strategy is shown as the pseudocode Algorithm \ref{algorithm1}. The basic idea of this strategy is to have it implemented in a greedy manner. At each step, we spend the least searches to determine the object’s latest location. Then, we iterate the step until we obtain the object’s current location, as the latest appearance is the most helpful for finding it at the current moment, and the amount of searches needed to find it again after its previous appearance will increase as the timespan becomes larger. Therefore, an indicator (see Equation (\ref{eq11})) is defined as the ratio of the needed amount (estimated by Equation (\ref{eq7})) to the timespan. At each step, we search the locations in descending order at a heuristic moment to minimize the indicator, as shown in lines 7 and 8 of the pseudocode. 

\begin{small}
	\begin{equation}
	\begin{split}
	{\mathop{\rm indicator}\nolimits} ({t_k}) = \frac{{en(\mathop {p'}\limits^ \to  ({c_x},{t_k}))}}{{{t_k} - {t_{cur}}}}
	\end{split}
	\label{eq11}
	\end{equation}
\end{small}

%算法一
\begin{algorithm}[h]
	\caption{: IHMs of Spatiotemporal Searches} %算法的名字
	\label{algorithm1}
	\hspace*{0in} \textbf{Input:} %算法的输入
	$c_x$, $d_x$, $t_x$, $l_{{t_p}}$, $t_p$, $TR'$ before $d_x$, $s({c_x},{l_i},{t_k})$ in $d_x$\\
	\hspace*{0in} \textbf{Output:} %算法的结果输出
	${l_{cur}}$, $amount$
	\begin{algorithmic}[1]
		\State initiate ${l_{cur}}=l_{{t_p}}$, ${t_{cur}}=t_p$, $amount = 0$;  % \State 后写一般语句
		\State partially known $tr({c_x},{d_x}) = \emptyset $;
		\While{${t_{cur}}<t_x$} % While语句，需要和EndWhile对应
		\State $tr({c_x},{d_x}) = tr({c_x},{d_x}) \cup \{ ({l_{cur}},{t_{cur}})\} $; // add the currently known location into the object’s trajectory
		\ForAll{${t_k} = ({t_{cur}} + 1)$ : $t_x$} // traverse all remaining moments
		\State get $\mathop {p'}\limits^ \to  ({c_x},{t_k})$ from $TR'$ and $tr({c_x},{d_x})$ with mobility prediction algorithm;
		\State ${t_{opt}} = \mathop {\arg \min }\limits_{{t_k} \in ({t_{cur}},{t_x}]} \frac{{en(\mathop {p'}\limits^ \to  ({c_x},{t_k}))}}{{{t_k} - {t_{cur}}}}$; // choose the moment which has the minimal indicator
		\State search locations at ${t_{opt}}$ according to $\mathop {p'}\limits^ \to  ({c_x},{t_{opt}})$ until $s({c_x},l{'_n}{t_{opt}}) = 1$; // search at the moment
		\State $l_{cur} = l{'_n}$, ${t_{cur}} = {t_{opt}}$, $amount = amount + n$; // update the current location and accumulate the cost
		\EndFor
		\EndWhile	
	\end{algorithmic}
\end{algorithm}

The model can be trained offline. The time complexity of the offline training is $O({n^3})$, while the time complexity of the online testing is $O({n^2})$.

\subsubsection{Intermediate Searching at Heuristic Spatiotemporal Units (IHUs)}

The intermediate searching at heuristic spatiotemporal units (IHUs) strategy is shown as the pseudocode Algorithm \ref{algorithm2}. The basic idea of this strategy is to implement it in a greedy manner similar to the IHMs but with two different aspects. 1) At each step, we search at a heuristic spatiotemporal unit (instead of searching the locations in descending order at a heuristic moment) to minimize an indicator and update the mobility prediction result regardless if the object is found in this unit. 2) The indicator uses $1/\mathop p\limits^ \to  ({c_x},{t_k})[j]$ (instead of $en(\mathop {p'}\limits^ \to  ({c_x},{t_2}))$) to estimate the amount of searches needed to find $c_x$ at $({l_j},{t_k})$ in $d_x$, e.g., if the probability of finding an object is $1/3$, then we can expect to find the object by executing three searches.

\begin{small}
	\begin{equation}
	\begin{split}
	{\mathop{\rm indicator}\nolimits} ({l_j},{t_k}) = \frac{{1/\mathop p\limits^ \to  ({c_x},{t_k})[j]}}{{{t_k} - {t_{cur}}}}
	\end{split}
	\label{eq12}
	\end{equation}
\end{small}

%%算法二
\begin{algorithm}[h]
	\caption{: IHUs of Spatiotemporal Searches} %算法的名字
	\label{algorithm2}
	\hspace*{0in} \textbf{Input:} %算法的输入
	$c_x$, $d_x$, $t_x$, $l_{{t_p}}$, $t_p$, $TR'$ before $d_x$, $s({c_x},{l_i},{t_k})$ in $d_x$\\
	\hspace*{0in} \textbf{Output:} %算法的结果输出
	${l_{cur}}$, $amount$
	\begin{algorithmic}[1]
		\State initiate ${l_{cur}}=l_{{t_p}}$, ${t_{cur}}=t_p$, $amount = 0$;  % \State 后写一般语句
		\State partially known $tr({c_x},{d_x}) = \emptyset $;
		\While{${t_{cur}}<t_x$} % While语句，需要和EndWhile对应
		\State $tr({c_x},{d_x}) = tr({c_x},{d_x}) \cup \{ ({l_{cur}},{t_{cur}})\} $; // add the currently known location into the object’s trajectory
		\ForAll{${t_k} = ({t_{cur}} + 1)$ : $t_x$} // traverse all remaining moments
		\State get $\mathop {p'}\limits^ \to  ({c_x},{t_k})$ from $TR'$ and $tr({c_x},{d_x})$ with mobility prediction algorithm;
		\State $({l_{opt}},{t_{opt}}) = \mathop {\arg \min }\limits_{{l_j} \in L,{t_k} \in ({t_{cur}},{t_x}]} \frac{{1/\mathop p\limits^ \to  ({c_x},{t_k})[j]}}{{{t_k} - {t_{cur}}}}$; // choose the spatiotemporal unit which has the minimal indicator
		\State search at the spatiotemporal unit $({l_{opt}},{t_{opt}})$ to get $s({c_x},{l_{opt}},{t_{opt}})$;
		\State $amount = amount + 1$; // accumulate the cost
		\If  $\ s({c_x},{l_{opt}},{t_{opt}}) = 1$ // find the object
		\State ${l_{cur}} = {l_{opt}}$, ${t_{cur}} = {t_{opt}}$; // update the current location
		\Else\ //$s({c_x},{l_{opt}},{t_{opt}}) = 0$, i.e., not find the object
		\State $TR' = TR' - \{ tr()\ {\mathop{\rm with}\nolimits}\  ({l_{opt}},{t_{opt}})\} $; // filter the history trajectories used for mobility prediction
		\EndIf
		\EndFor
		\EndWhile	
	\end{algorithmic}
\end{algorithm}

The model cannot be trained offline. We need to rematch the training data after each search has been completed. The time complexity of its online training and testing is $O({n^3})$.

\subsection{Comparative Explanation}

\begin{figure}[h]
	\centering
	\begin{minipage}[t]{8cm}
		\vspace{-0.1cm} %设置与上面正文的距离
		\setlength{\abovecaptionskip}{0.2cm}   %调整图片标题与图距离
		\setlength{\belowcaptionskip}{-0.2cm}   %调整图片标题与下文距离
		%		\subfigcapskip=-0.3cm %设置子图与子标题之间的距离
		\centering 
		\includegraphics[width=8cm]{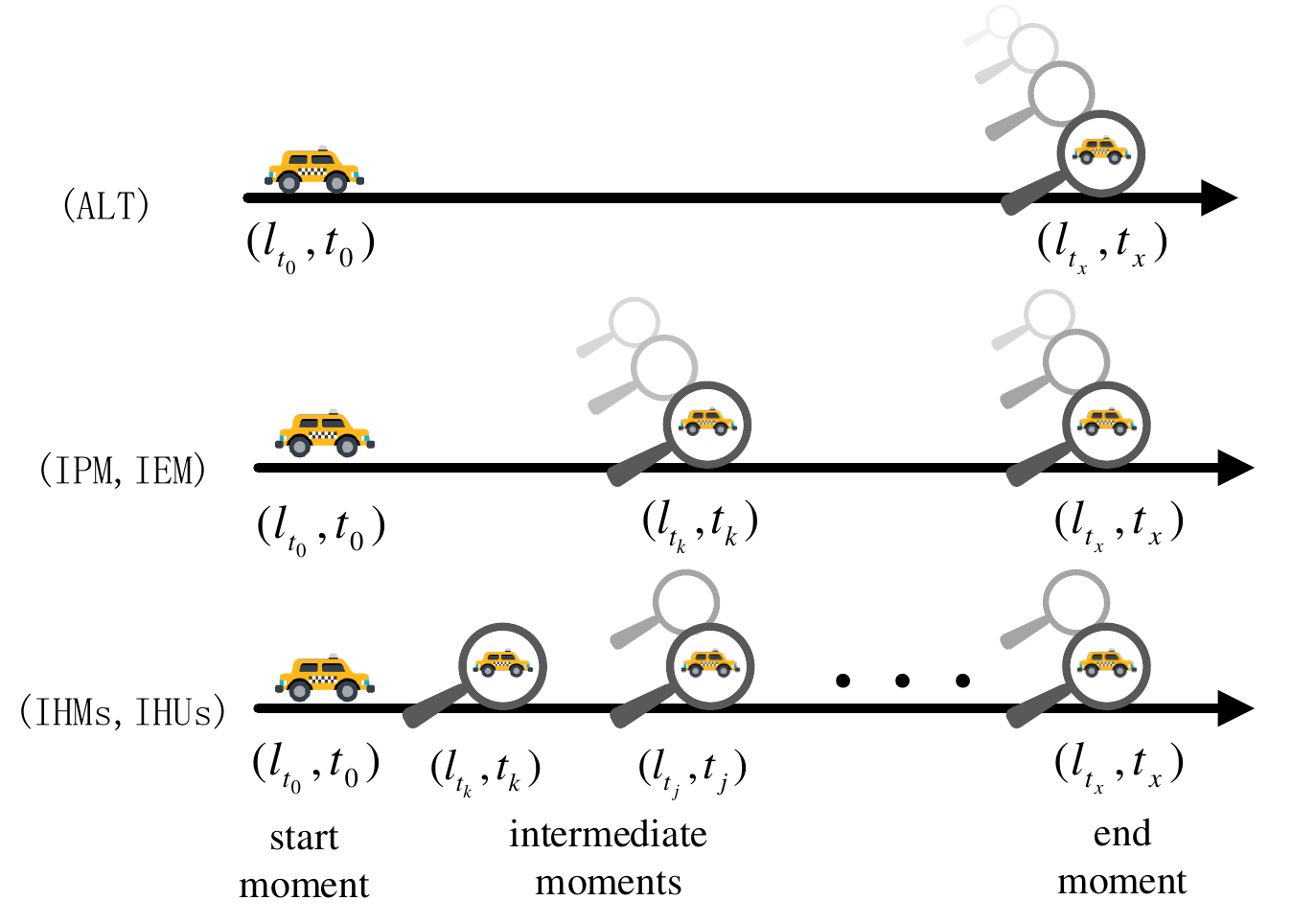}
		\caption{A comparative diagram of five searching strategies}
		\label{figure3}
	\end{minipage}
\end{figure}

We provide comparative explanation of the above five searching strategies to make them easily understood. As shown in Fig. \ref{figure3}, ALT directly searches the object at the end moment  to find it (no intermediate search). IPM and IEM first search at an intermediate moment, and then search at the end moment  (one intermediate search). IHMs and IHUs seek the object at intermediate moments, until finally find it at the end moment  (adaptive intermediate searches). The difference between IPM and IEM: the intermediate moment is specified differently. The difference between IHMs and IHUs: IHMs first determines an intermediate moment and then searches from locations with higher probability to lower probability, while IHUs determines an intermediate moment and corresponding location at the same time.

\section{Evaluation}
\label{section4}

\subsection{Dataset Description and Preprocessing}

\begin{figure}[htbp]
	\vspace{-0.3cm} %设置与上面正文的距离
	\setlength{\abovecaptionskip}{0.2cm}   %调整图片标题与图距离
	\setlength{\belowcaptionskip}{-0.2cm}   %调整图片标题与下文距离
	\subfigcapskip=0cm %设置子图与子标题之间的距离
	\centering
	\subfigure[Raw trajectory distribution]{%子图名字为空
		\begin{minipage}[t]{0.45\linewidth}
			\centering
			\includegraphics[width=4cm]{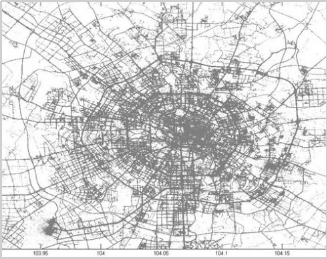}
			\label{figure4a}%添加标签方便引用
		\end{minipage}%
	}%
	\subfigure[Raw trajectory example]{
		\begin{minipage}[t]{0.55\linewidth}
			\centering
			\includegraphics[width=5cm]{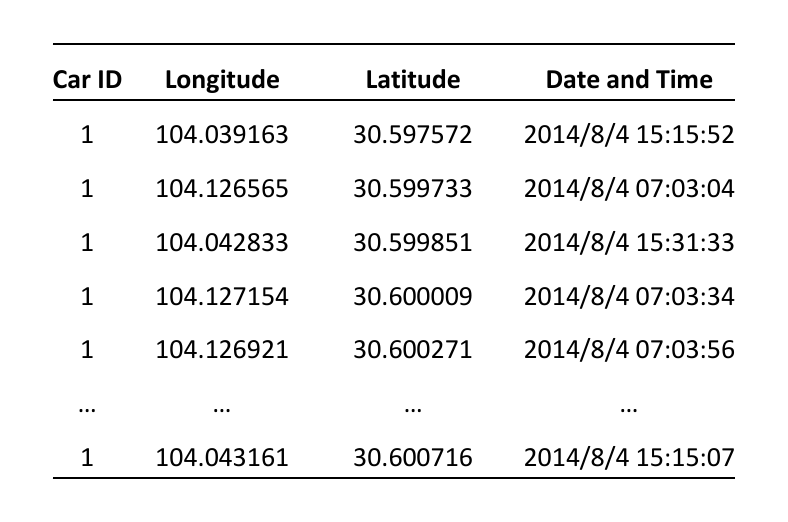}
			\label{figure4b}
		\end{minipage}%
	}%
	\centering
	\caption{Raw trajectories with GPS waypoints}
	\label{figure4}
\end{figure}

\begin{figure}[htbp]
	\vspace{-0.3cm} %设置与上面正文的距离
	\setlength{\abovecaptionskip}{0.2cm}   %调整图片标题与图距离
	\setlength{\belowcaptionskip}{-0.2cm}   %调整图片标题与下文距离
	\subfigcapskip=0cm %设置子图与子标题之间的距离
	\centering
	\subfigure[Discretized trajectory distribution]{%子图名字为空
		\begin{minipage}[t]{0.48\linewidth}
			\centering
			\includegraphics[width=4cm]{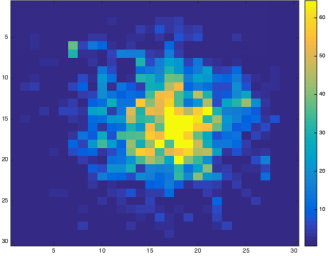}
			\label{figure5a}%添加标签方便引用
		\end{minipage}%
	}%
	\subfigure[Discretized trajectory example]{
		\begin{minipage}[t]{0.48\linewidth}
			\centering
			\includegraphics[width=4cm]{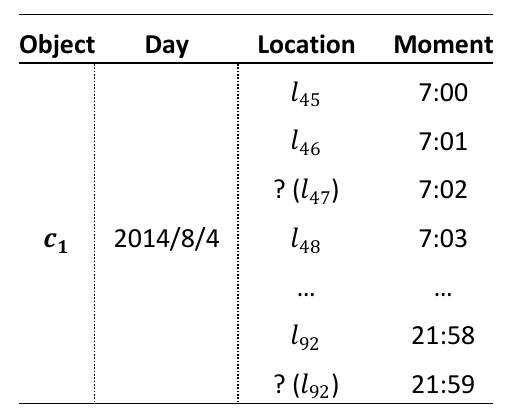}
			\label{figure5b}
		\end{minipage}%
	}%
	\centering
	\caption{Discretized and complemented trajectories}
	\label{figure5}
\end{figure}

The dataset had been derived from the trajectories of 19,000 taxis in Chengdu, China in August 2014. A rectangular urban area of approximately 30 km × 30 km is selected, and the time is from 7:00 to 21:59 over 17 days, resulting in a scoped and denser dataset. A raw trajectory of a day of a car is composed of thousands of GPS waypoints, each representing a car’s identification, latitude and longitude, and date and time, as shown in Fig. \ref{figure4}. As mentioned previously, the urban area is discretized into locations, each covering approximately 1 km × 1 km (900 locations in total), and the time is discretized into moments, each one lasting 1 minute (900 moments in total). Then, the raw trajectories are discretized, as shown in Fig. \ref{figure5}. When a car passes through several locations in a minute, we only preserve the first location of that minute for simplification.

We assume the objects in the same dataset to have the same spatiotemporal characteristics. In this manner, the normal history trajectories can help to predict the mobility of a target object. Objects with abnormal trajectories may be hard to search (i.e., with a much larger amount of searches) but can still be searched. To this end, using taxi traces is justifiable for the experiments. Note that in real application, the police may not be able to locate the car by GPS or a direct phone call because of the following reasons: the car does not have a GPS receiver; the car’s GPS signal does not upload into the police system; the police does not know who are in or around the car and which phone numbers can be reached; or the police has decided not to arouse suspicion. Only massive camera records are available for spatiotemporal searching.

A preprocessing operation worth mentioning is on how to deal with missing data, i.e., certain time-specific locations unknown in a trajectory, because of GPS signal loss or the object is out of scope. The two identified situations are as follows: 1) the unknown locations are at the start or end of the trajectory, and 2) the unknown locations are between two known locations. For the first situation, if the time with unknown locations lasts less or equal than five moments, then the car is deemed at the same location as the recent moment with a known location; otherwise, the trajectory is discarded. For the second situation, if the time with unknown locations lasts less or equal than ten moments, and the distance of two known locations are no longer than a 15 km threshold, then the car’s locations are interpolated according to a uniform rectilinear movement; otherwise, the trajectory is discarded. Approximately 70\% trajectories had been retained. Among them, 0.3\% are missing time-specific locations but can be complemented.

%并排插入图片

%并排插入图片

The 99,265 trajectories of the former 14 days are used as the training data, i.e., $TR'$. The 21,963 trajectories of latter 3 days are used as the testing data. In running the test, a trajectory is picked from testing data as the ground truth of $tr({c_x},{d_x})$. Then, in each experiment, we exhaustively run 21,963 tests and average the results.

\subsection{Performance of Mobility Prediction}

Usually, the Top-N accuracy is adopted as the performance metric of mobility prediction algorithms. However, in this study, the objective of the mobility prediction is to guide the least spatiotemporal search and estimate the amount of required searches. The Top-N accuracy is not a proper metric for this objective. For instance, given two mobility prediction results, such as $\left\langle {0.5({l_1}),0.3({l_2}),0.2({l_3})} \right\rangle $ and $\left\langle {0.5({l_1}),0.1({l_2}),0.4({l_3})} \right\rangle $, and the ground truth $l_3$, their Top-1 accuracy are the same (i.e., 0), but the amounts of searches are different (i.e., three and two, respectively). Therefore, we directly choose the amount of searches that had been executed as the metric of the mobility prediction algorithms.

Beside the second-order Markov model that may outperform the first-order Markov model, another plausible improvement is the time-specific first-order Markov model, whose $TPM$ is trained only from the trajectories with the same specified start moment, as depicted in Equations (\ref{eq13}) and (\ref{eq14}). Likewise, the time-specific second-order Markov model is also tested.

\begin{small}
	\begin{equation}
	\begin{split}
	TPM_{\Delta t,{t_p}}^{|L| \times |L|} = \left\{ {p({l_i},{l_j})|\Delta t,{t_p}} \right\}
	\end{split}
	\label{eq13}
	\end{equation}
\end{small}

\begin{small}
	\begin{equation}
	\begin{split}
	p({l_i},{l_j})|\Delta t,{t_p} = \frac{{\# tr(){\mathop{\rm with}\nolimits} ({l_i},{t_p})\& ({l_j},{t_p} + \Delta t)}}{{\# tr(){\mathop{\rm with}\nolimits} ({l_i},{t_p})}},tr() \in TR'
	\end{split}
	\label{eq14}
	\end{equation}
\end{small}

\begin{figure}[htbp]
	\vspace{-0.3cm} %设置与上面正文的距离
	\setlength{\abovecaptionskip}{0.2cm}   %调整图片标题与图距离
	\setlength{\belowcaptionskip}{-0.2cm}   %调整图片标题与下文距离
	\subfigcapskip=0cm %设置子图与子标题之间的距离
	\centering
	\subfigure[Fixed start moment]{%子图名字为空
		\begin{minipage}[t]{0.48\linewidth}
			\centering
			\includegraphics[width=4.1cm]{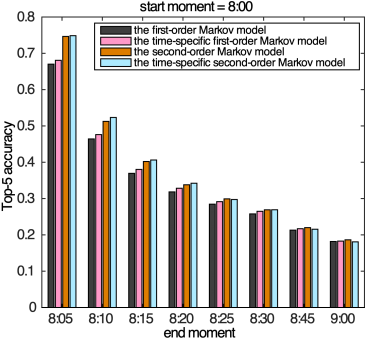}
			\label{figure6a}%添加标签方便引用
		\end{minipage}%
	}%
	\subfigure[Fixed end moment]{
		\begin{minipage}[t]{0.48\linewidth}
			\centering
			\includegraphics[width=4.1cm]{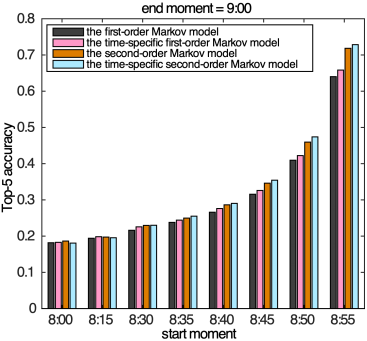}
			\label{figure6b}
		\end{minipage}%
	}%
	
	\subfigure[Fixed start moment]{
		\begin{minipage}[t]{0.48\linewidth}
			\centering
			\includegraphics[width=4.1cm]{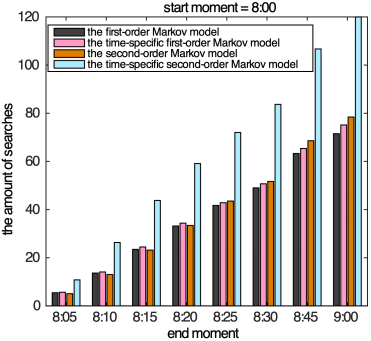}
			\label{figure6c}
		\end{minipage}%
	}%
	\subfigure[Fixed end moment]{
		\begin{minipage}[t]{0.48\linewidth}
			\centering
			\includegraphics[width=4.1cm]{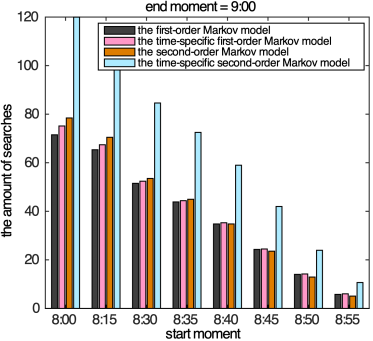}
			\label{figure6d}
		\end{minipage}%
	}%
	\centering
	\caption{Different performance metrics of mobility prediction algorithms. (a),(b) Top-5 accuracy (higher is better); (c),(d) amount of searches (lower is better).}
	\label{figure6}
\end{figure}

As shown in Fig. \ref{figure6}, although the other three models perform better than the first-order Markov model in terms of the Top-5 accuracy, the first-order Markov model outperforms the other three models in terms of amount of searches. That is to say, the correlation between Top-N accuracy and amount of searches is weak. Fundamentally, Top-N accuracy is a metric of classification, whereas the amount of searches is a metric of ranking. If the performance is measured with the amount of searches, then its impact on the total cost is straightforward, i.e., a better mobility prediction algorithm will result in lower total cost.

%并排插入图片

\begin{figure}[ht]
	\centering
	\begin{minipage}[t]{9cm}
		\vspace{-0.2cm} %设置与上面正文的距离
		\setlength{\abovecaptionskip}{0.2cm}   %调整图片标题与图距离Fig. \ref{figure1}
		\setlength{\belowcaptionskip}{-0.15cm}   %调整图片标题与下文距离
		%		\subfigcapskip=-0.3cm %设置子图与子标题之间的距离
		\centering 
		\includegraphics[width=9cm]{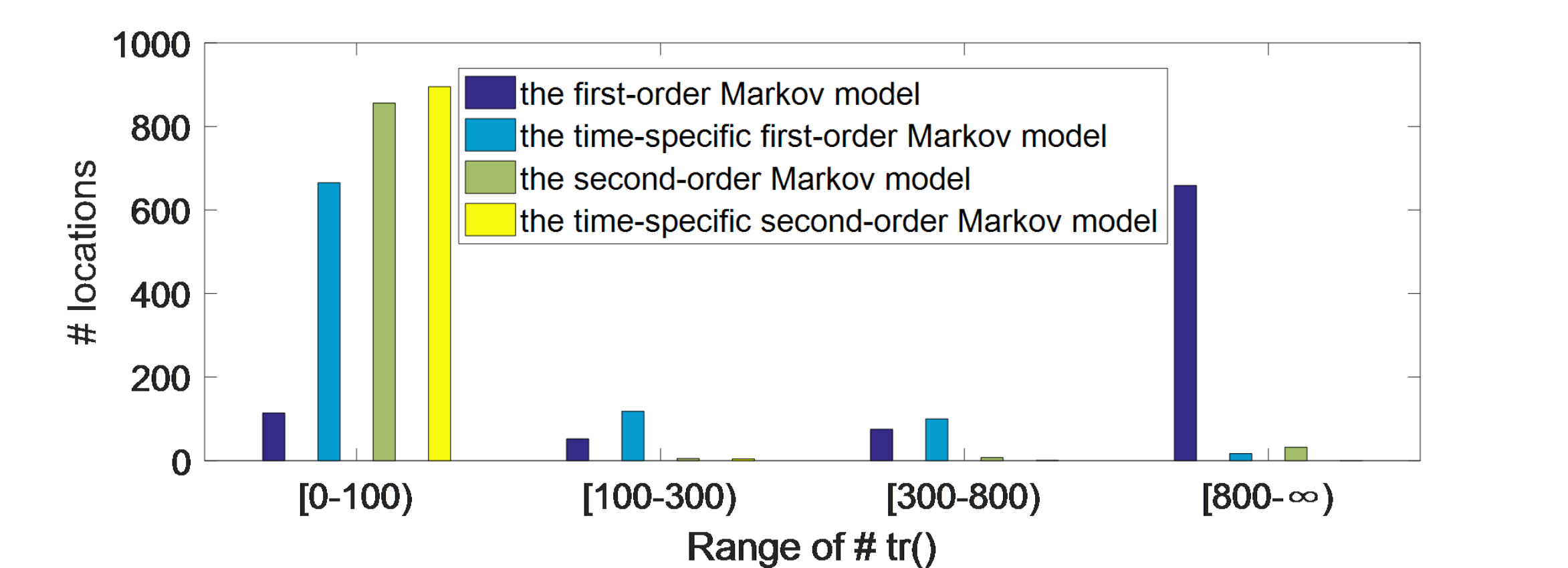}
		\caption{Statistics of trajectories passing through different locations. The first deep blue bar means that approximately 100 locations (total of 900 locations) have passed through 0 to 100 trajectories, etc. For the second-order Markov models, the first location is fixed to an extremely busy location l495 and the moment of the second locations is 5 minutes later. For the time-specific models, the moment is fixed to 8:00.}
		\label{figure7}
	\end{minipage}
\end{figure}

The number of trajectories passing through the given locations reflects the amount of data that can be used to train a mobility prediction model. Fig. \ref{figure7} shows the statistics of the trajectories passing through different locations in the training data. For the mobility prediction based on the first-order Markov model, most locations have plenty of matched history trajectories (approximately 1,000 on the average); by contrast, for the other three models, most locations have few trajectories (by the dozens on the average). Therefore, as the matched history trajectories for training are insufficient, the time-specific models or the high-order Markov models will be inaccurate because of low SNR or overfitting.

\begin{figure}[ht]
	\centering
	\begin{minipage}[t]{9cm}
		\vspace{-0.2cm} %设置与上面正文的距离
		\setlength{\abovecaptionskip}{0.2cm}   %调整图片标题与图距离Fig. \ref{figure1}
		\setlength{\belowcaptionskip}{-0.15cm}   %调整图片标题与下文距离
		%		\subfigcapskip=-0.3cm %设置子图与子标题之间的距离
		\centering 
		\includegraphics[width=7cm]{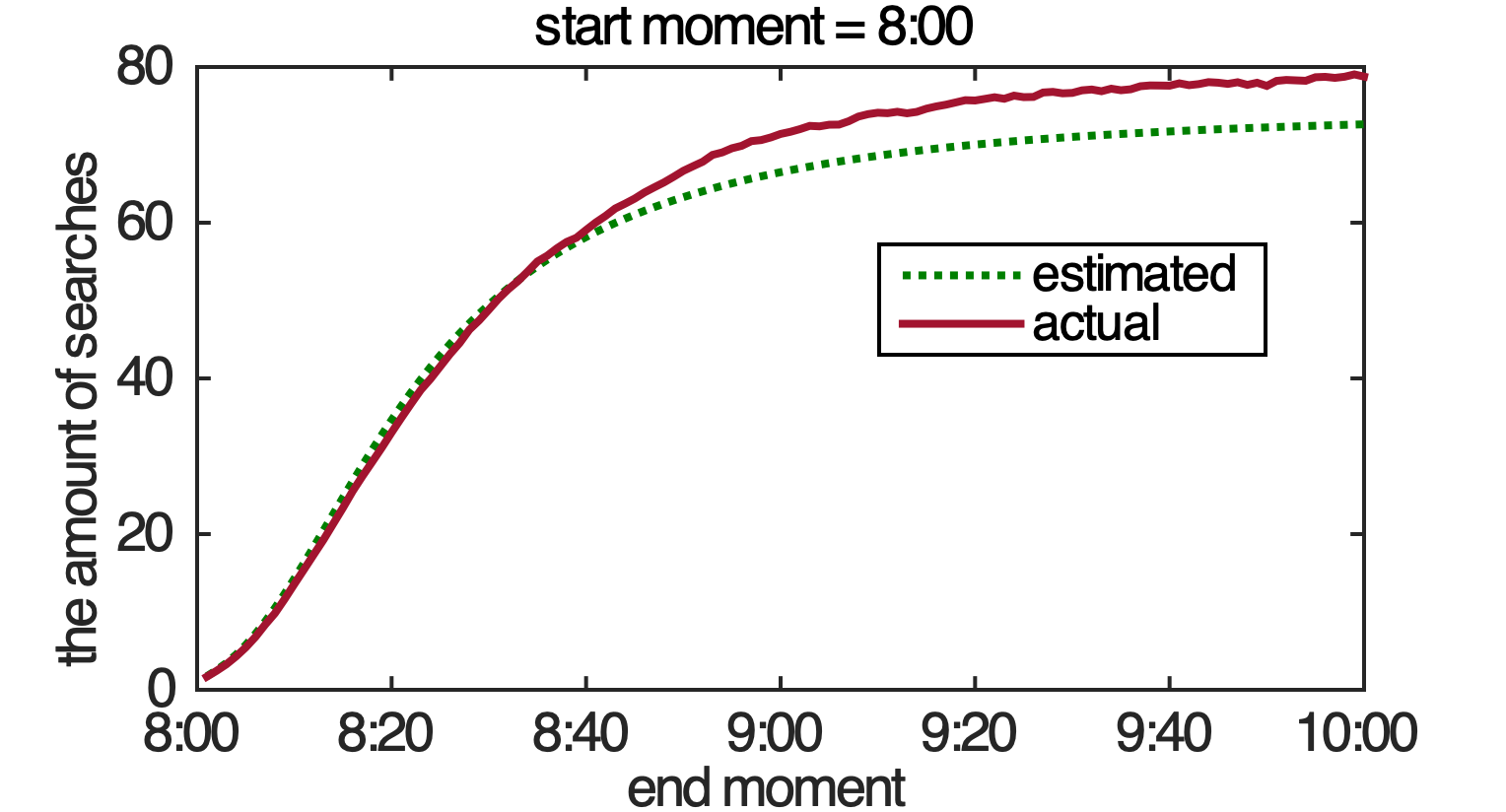}
		\caption{Amount of searches estimated by the first-order Markov model}
		\label{figure8}
	\end{minipage}
\end{figure}

The mobility prediction algorithm is also used to estimate the amount of required searches in the IEM or IHMs strategies. Here, we show how good the estimation is provided by the first-order Markov model (Fig. \ref{figure8}). When the transition time (i.e., ${\Delta t}$ = end moment – start moment) is shorter than 35 minutes, the estimated amount of searches almost coincides with the actual amount. However, when the transition time becomes longer, the estimated amount obviously becomes smaller than the actual amount. This phenomenon means that the effectiveness of the first-order Markov model is attenuated towards long-term mobility prediction. As a consequence, the strategies relying on this approach may perform worse. The scenario in Fig. \ref{figure11} also verifies this observation.

\begin{figure}[htbp]
	\vspace{-0.3cm} %设置与上面正文的距离
	\setlength{\abovecaptionskip}{0.2cm}   %调整图片标题与图距离
	\setlength{\belowcaptionskip}{-0.2cm}   %调整图片标题与下文距离
	\subfigcapskip=-0.4cm %设置子图与子标题之间的距离
	\centering
	\subfigure[Transition time is 5min]{%子图名字为空
		\begin{minipage}[t]{0.5\linewidth}
			\centering
			\includegraphics[width=4.6cm]{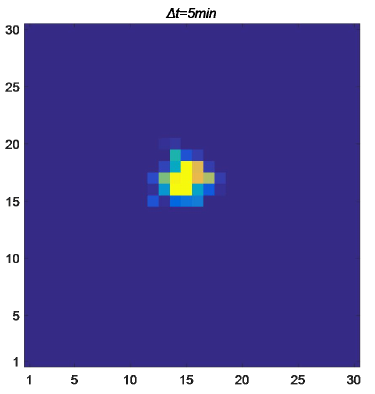}
			\label{OverView figure111}%添加标签方便引用
		\end{minipage}%
	}%
	\subfigure[Transition time is 30min]{
		\begin{minipage}[t]{0.5\linewidth}
			\centering
			\includegraphics[width=4.6cm]{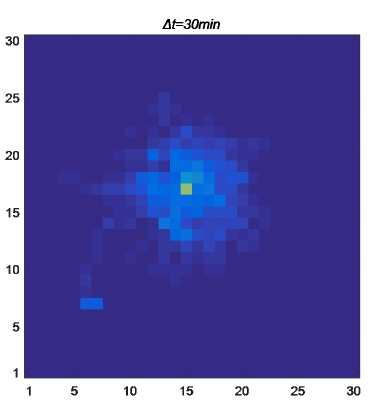}
			\label{figure112}
		\end{minipage}%
	}%
	
	\subfigure[Transition time is 60min]{
		\begin{minipage}[t]{0.5\linewidth}
			\centering
			\includegraphics[width=5cm]{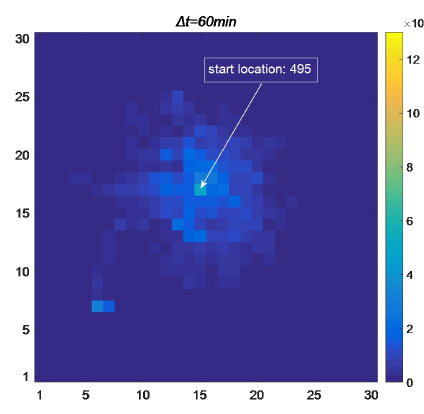}
			\label{figure113}
		\end{minipage}%
	}%
	\centering
	\caption{Distribution of end locations after different transition time
		(start location: ${l_{495}}$).}
	\label{figure9}
\end{figure}

The attenuation of the first-order Markov model at the longer transition time can be explained by the predictability of trajectories. Fig. \ref{figure9} shows the distribution of the end locations after different transition time (the start location is fixed to a very busy location ${l_{495}}$). The distribution of the end location is clearly dispersed after a longer transition time. This scenario also further worsens the data sparsity problem.

%并排插入图片

\subsection{Results of Different Searching Strategies}

The five searching strategies that described in Section \ref{section3}, i.e., the ALT, IPM, IEM, IHMs, and IHUs strategies, are compared under different experimental settings. 1) The ${\Delta t}$ of the first setting is fixed to 30 minutes, and the start moment (i.e., past moment) is assigned to 7:00–21:30 at 1-minute intervals. 2) The start moment of the second setting is fixed to 8:00, and the end moment (i.e., current moment) is assigned to 8:05–9:20 at 5-minute intervals. 3) The start and end moments of the third setting is fixed to 8:00 and 8:30, respectively, but the start locations are divided into four groups according to their busyness. As the IPM strategy needs to specify a parameter beforehand; here, we simply specify it right in the middle of the time, i.e., ${t_p} + \Delta t/2$. The variable parameter is discussed in Subsection \ref{subsection45}.

\begin{figure*}[ht]
	\centering
	\begin{minipage}[t]{16cm}
		\setlength{\abovecaptionskip}{0.2cm}   %调整图片标题与图距离Fig. \ref{figure1}
		\setlength{\belowcaptionskip}{-0.25cm}   %调整图片标题与下文距离
		%		\subfigcapskip=-0.3cm %设置子图与子标题之间的距离
		\centering 
		\includegraphics[width=12cm]{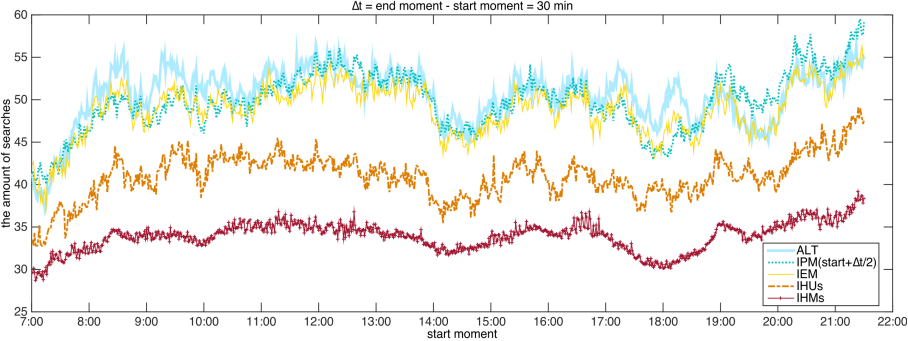}
		\caption{ Results of different searching strategies (fix $\Delta t$).}
		\label{figure10}
	\end{minipage}
\end{figure*}

\begin{figure*}[ht]
	\centering
	\begin{minipage}[t]{16cm}
		\setlength{\abovecaptionskip}{0.2cm}   %调整图片标题与图距离Fig. \ref{figure1}
		\setlength{\belowcaptionskip}{-0.25cm}   %调整图片标题与下文距离
		%		\subfigcapskip=-0.3cm %设置子图与子标题之间的距离
		\centering 
		\includegraphics[width=12cm]{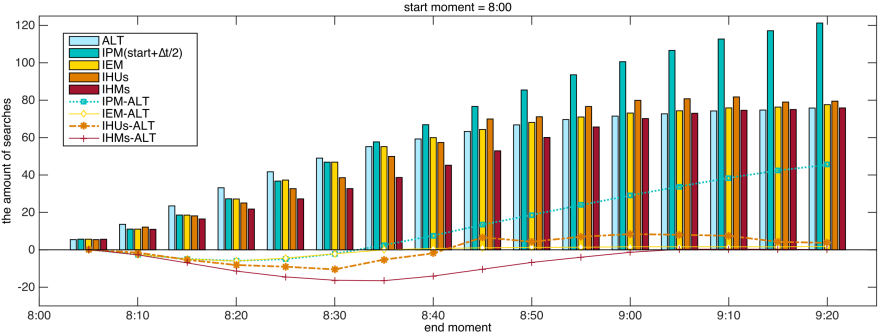}
		\caption{Results of different searching strategies (fix ${t_p}$).}
		\label{figure11}
	\end{minipage}
\end{figure*}

1) The result of the first setting is shown as Fig. \ref{figure10}. The IHMs strategy outperforms the other strategies at all start moments. The second best strategy is the IHUs. The bottom three, i.e., ALT, IPM, and IEM, are almost equivalent in terms of amount of searches, requiring approximately 33, 40, and 50 spatiotemporal searches to find a car that had disappeared at half an hour earlier. We can also determine the different start moments influencing the searching difficulty, e.g., the easiest start moments are near 7:00, 14:10, or 18:00, whereas the hardest start moments are near 8:30, 12:00, 16:30, or 21:30. Thus, track a car during rush hours is easier in general, which can be attributed to higher mobility predictability caused by the low speed during rush hours.

2) The result of the second setting is shown as Fig. \ref{figure11}. Similar to the first setting, the IHMs strategy outperforms the other strategies at all end moments. A longer $\Delta t$ raises the costs (i.e., the amount of searches) to find a car, which is consistent with common sense. The superiority of IHMs is most prominent when $\Delta t$ = 35 minutes. When $\Delta t <$ 35 minutes, the order of rank of the cost is IHMs $<$ IHUs $<$ IPM $ \approx $ IEM $<$ ALT. When 35 minutes $ \le \Delta t$ $<$ 60 minutes, the order of rank is IHMs $<$ IEM $ \approx $ ALT $ \approx $ IHUs $<$ IPM. When $ \Delta t \ge$ 60 minutes, all strategies perform similarly except IPM. The trends reveal that the IPM strategy cannot adapt to different searching difficulties. ALT, as the simplest strategy, and the other strategies tend to perform better than the IPM, but it does not always hold. The polylines in Fig. \ref{figure10} show the difference between the other strategies and ALT, with other details discussed in Subsection \ref{subsection44}. Thus, for the question as to whether the costs spent in the earlier searching can be compensated by the costs saved in latter searching, the answer depends primarily on the searching difficulty and secondarily on the intermediate searching strategy. If a car is out of sight for a long time, then intermediate searching does not seem to help. Meanwhile, if a car is out of sight for a short time, then an efficient searching strategy can save up to 1/3 of the costs.

%并排插入图片
\begin{figure}[htbp]
	\vspace{-0.3cm} %设置与上面正文的距离
	\setlength{\abovecaptionskip}{0.2cm}   %调整图片标题与图距离
	\setlength{\belowcaptionskip}{-0.2cm}   %调整图片标题与下文距离
	\subfigcapskip=0cm %设置子图与子标题之间的距离
	\centering
	\subfigure[Different busyness of start location]{%子图名字为空
		\begin{minipage}[t]{0.42\linewidth}
			\centering
			\includegraphics[width=4cm]{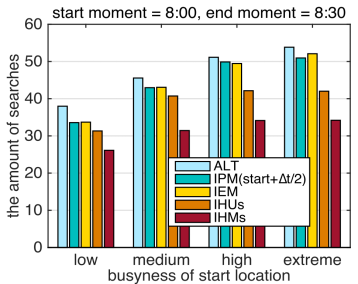}
			\label{figure12a}%添加标签方便引用
		\end{minipage}%
	}%
	\subfigure[Statistics of location busyness]{
		\begin{minipage}[t]{0.56\linewidth}
			\centering
			\includegraphics[width=5cm]{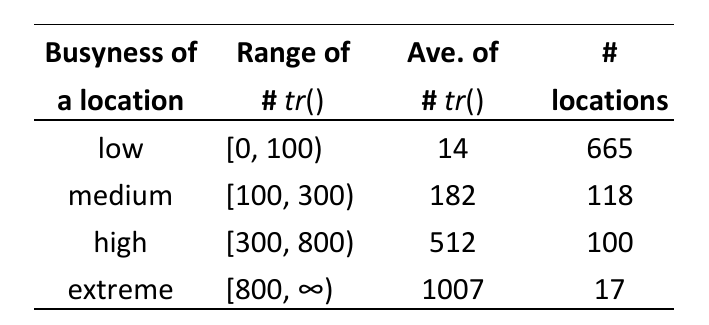}
			\label{figure12b}
		\end{minipage}%
	}%
	\centering
	\caption{Results of different searching strategies (fix $t_p$ and ${\Delta t}$)}
	\label{figure12}
\end{figure}

3) The result of the third setting is shown as Fig. \ref{figure12a}. The IHMs strategy still outperforms the other strategies at all start locations. The busyness of a location is characterized by how many trajectories pass through it. We divide all 900 locations into four groups according to their busyness, as shown in Fig. \ref{figure12b}. We find that the busier start location leads to more costs in tracking a car.

As the IHMs strategy outperforms the other strategies in all three settings, we propose the use of IHMs in this study. The following subsections provide detailed comparisons of the IHMs strategy and the other strategies as a means of explaining why IHMs can save the highest amount of searches.

\subsection{Detailed Comparison of IHMs vs. ALT}
\label{subsection44}

\begin{figure}[ht]
	\centering
	\begin{minipage}[t]{9cm}
		\vspace{-0.2cm} %设置与上面正文的距离
		\setlength{\abovecaptionskip}{0.2cm}   %调整图片标题与图距离Fig. \ref{figure1}
		\setlength{\belowcaptionskip}{-0.15cm}   %调整图片标题与下文距离
		%		\subfigcapskip=-0.3cm %设置子图与子标题之间的距离
		\centering 
		\includegraphics[width=8cm]{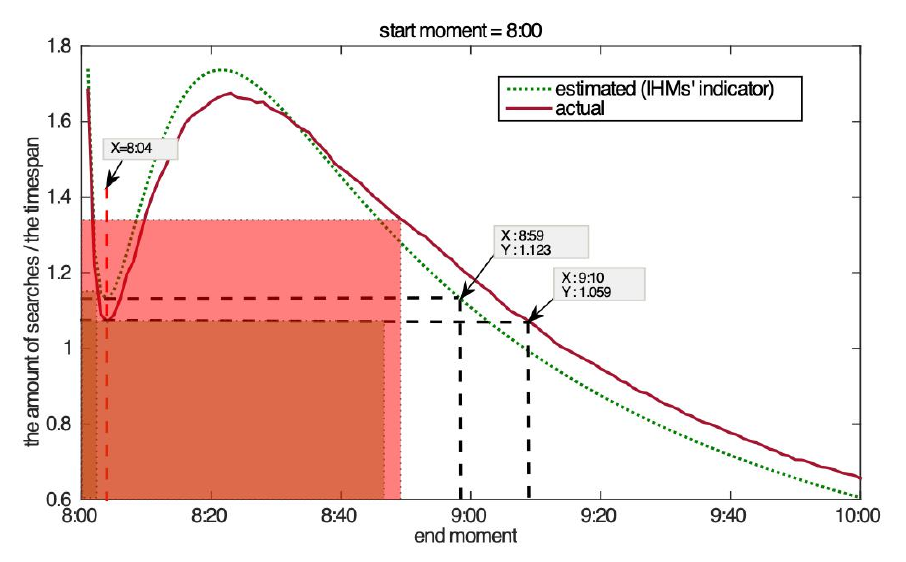}
		\caption{Trend of the IHMs’ indicator}
		\label{figure13}
	\end{minipage}
\end{figure}

As mentioned previously, the IHMs’ indicator is the ratio of the estimated amount of searches to the timespan (according to Equation (\ref{eq11})). The inclination angle of the green curve shown in Fig. \ref{figure8} represent this ratio, as shown in Fig. \ref{figure13}. The actual ratio in Fig. \ref{figure13} is the average of all testing trajectories. For a particular testing trajectory, the actual ratio adheres to the same trend but with small fluctuations. The ALT strategy always executes searches at the last time; the corresponding amount of searches is equal to the area of the light red rectangle shown in Fig. \ref{figure13}. By contrast, the IHMs strategy executes searches at the moment with the smallest estimated ratio; the corresponding amount of searches is equal to the area of the deep red rectangle shown in Fig. \ref{figure13}. According to the trend of the indicator, when ${\Delta t}$ is shorter than 4 minutes or longer than 60 minutes, the IHMs will be perform the same as ALT, as the last time has the smallest ratio; otherwise, IHMs will execute searches in multiple steps, in which each step will span approximately 4 minutes, as the time span of 4 minutes has the smallest ratio. The area difference between the deep red rectangle and the light red rectangle is the largest when ${\Delta t}$ = 35 minutes.

\subsection{Detailed Comparison of IHMs vs. IPM and IEM}
\label{subsection45}
\begin{figure}[ht]
	\centering
	\begin{minipage}[t]{9cm}
		\vspace{-0.2cm} %设置与上面正文的距离
		\setlength{\abovecaptionskip}{0.2cm}   %调整图片标题与图距离Fig. \ref{figure1}
		\setlength{\belowcaptionskip}{-0.15cm}   %调整图片标题与下文距离
		%		\subfigcapskip=-0.3cm %设置子图与子标题之间的距离
		\centering 
		\includegraphics[width=9cm]{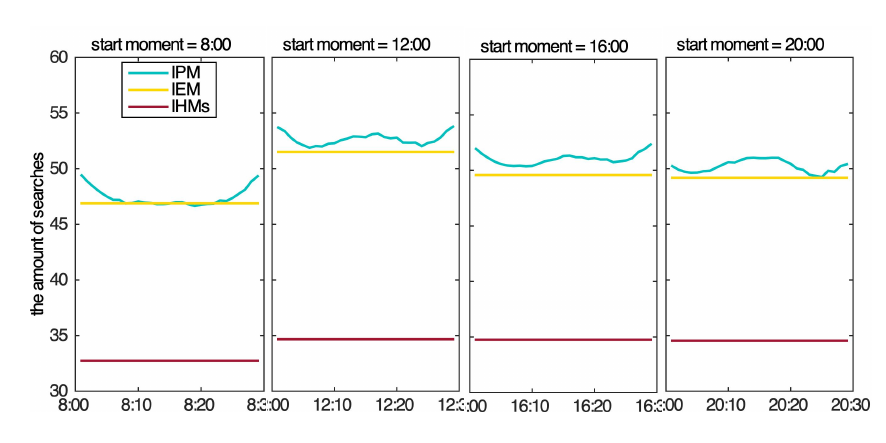}
		\caption{Results of IPM (variable parameter), IEM, and IHMs}
		\label{figure14}
	\end{minipage}
\end{figure}

According to the above discussion, if ${\Delta t}$ is between 4 minutes and 60 minutes, we can benefit from searches at intermediate moments. Here, we fix ${\Delta t}$ to 30 minutes and set the start moment to 8:00, 12:00, 16:00, and 20:00. Then, we compare the three strategies with searches at the intermediate moment/s, i.e., IHMs, IPM, and IEM strategies. Fig. \ref{figure14} shows the comparative results. The parameter of IPM traverses all moments from $t_p$ to $t_x$. Different intermediate searching moments result in varying performances, in which the best performance is reached at $t_p + 4$ minutes or $t_p - 4$ minutes. This finding can be attributed to the 4 minutes having the lowest cost-timespan ratio. If only one intermediate moment is available, then the IEM can find it correctly, which is the moment that allows IPM to reach its lower bound. The IHMs outperforms the other strategies because it can exploit approximately 30 minutes/4 minutes $ \approx $ 7 of intermediate moments, whereas the other two strategies only can exploit one intermediate moment.

\subsection{Detailed Comparison of IHMs vs. IHUs}

\begin{figure}[ht]
	\centering
	\begin{minipage}[t]{9cm}
		\vspace{-0.2cm} %设置与上面正文的距离
		\setlength{\abovecaptionskip}{0.2cm}   %调整图片标题与图距离Fig. \ref{figure1}
		\setlength{\belowcaptionskip}{-0.15cm}   %调整图片标题与下文距离
		%		\subfigcapskip=-0.3cm %设置子图与子标题之间的距离
		\centering 
		\includegraphics[width=5cm]{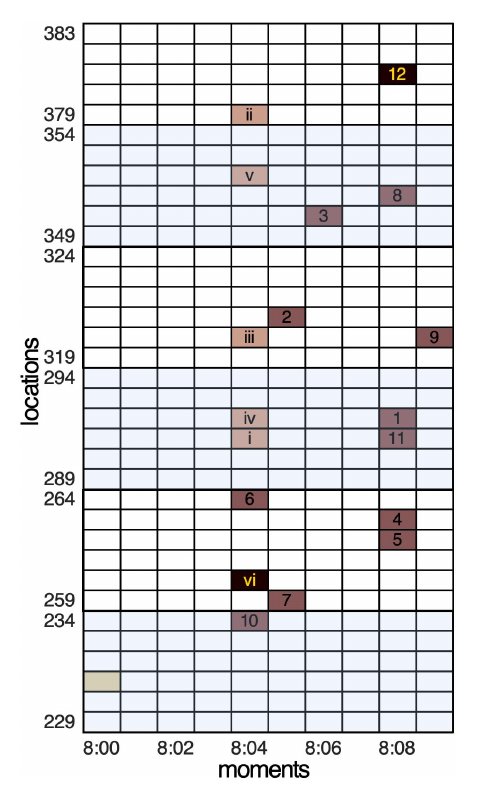}
		\caption{Order of searched units by IHMs and IHUs for a special case. The car was at (8:00, $l_{231}$). The IHMs spend six searches (i, ii, …) to find the car at (8:04, $l_{260}$), whereas the IHUs spend 12 searches (1, 2, …) to find the car at (8:08, $l_{381}$).}
		\label{figure15}
	\end{minipage}
\end{figure}

The IHUs had not performed as good as IHMs, which is somewhat unexpected. The IHUs attempted to determine which spatiotemporal unit is the best to search at each next-step, whereas the IHMs attempted to determine which moment is the best one to search each next-step. However, if we explore deeper into their searching processes, we find the following reasons: 1) as an element in $\mathop p\limits^ \to  ({c_x},{t_k})$ is only comparable within the specific moment $t_k$, it is senseless to use $1/\mathop p\limits^ \to  ({c_x},{t_k})[i]$ to estimate and compare the amount of searches among different moments; and 2) staying in the same moment to search locations one by one can fully utilize the information gained from the previous failed searches. In other words, search a location at a moment increases the value of exploring other locations at that moment, when the object was not found by that search. For example, at a moment $t_k$ with appearing probabilities $\left\langle {0.5({l_1}),0.3({l_2}),0.2({l_3})} \right\rangle $, the first search at $({l_1},{t_k})$ fails can inform us about $t_k$ with $\left\langle {0({l_1}),0.6({l_2}),0.4({l_3})} \right\rangle $, but the information about the other moments are still lacking. Fig. \ref{figure15} shows the order of searched units by the IHMs and IHUs for a special case. On the average, 1/4 of the amount of searches by IHMs are hit, whereas only 1/8 of the amount of searches by the IHUs are hit. Obviously, the hit searches can offer more information than failed ones, hence the reason why the IHMs are more efficient than the IHUs.

\section{Conclusion}
\label{section6}

\subsection{Summary}

Tracking a car or a person in a city is crucial in urban safety management. How can we complete the task with a minimal number of spatiotemporal searches? This study solves the problem by designing and comparing five searching strategies and validating that the proposed one, which is named the IHMs strategy, is the most efficient. The basic idea of the IHMs strategy is to execute it in a greedy manner. At each step, we determine which moment is the best one to search according to a heuristic indicator. Then, at each moment, the locations are searched one by one in descending order of predicted appearing probabilities until a search is hit. We iterate this step until we are able to determine the object’s current location. The heuristic indicator represents the ratio of the estimated amount of needed searches for finding the object at the considered moment to the timespan from the previous moment to the considered moment, also called the cost-timespan ratio for brevity. The cost-timespan ratio can resolve the conflict of spending the least searches to derive the object’s latest location.

IHMs is one of the strategies with intermediate searching, but not all intermediate searching strategies can outperform the simple ALT all the time. For the question as to whether the costs spent in earlier searching can be compensated by the costs saved in latter searching, the answer depends primarily on the searching difficulty and secondarily on the intermediate searching strategy. Moreover, the time length of being out of sight is the most influential factor of spatiotemporal searching difficulty, along with other factors, such as mobility prediction effectiveness, and start moment and start location, e.g., tracking a car at a busy location during rush hours is more difficult. In general, if a car is out of sight for a long time ($ \ge $60 minutes), intermediate searching seems inappropriate. Meanwhile, if a car is out of sight for a short time, then an efficient searching strategy can save up to 1/3 of the costs.

In the process of spatiotemporal searches, the objective of the mobility prediction algorithm is to guide the least spatiotemporal search and estimate the amount of required searches. We propose that a proper metric for this objective is the amount of searches rather than the Top-N accuracy. With this new metric, the first-order Markov model can outperform its other three variants.

\subsection{Discussions}

Spatiotemporal searching allows us to search at the past moments, which means that we have to record initially the history of the past. Nowadays, video/image data are prevalent for recording in many applications. Although only camera records are taken as examples in this study, other records with location and moment indexes are also applicable. Thus, the strategy proposed in this study does not lose its generality. Cameras are multifarious, not only include those for traffic and public surveillance, but also those of citizens’ smart mobile phones. Thus, assuming that these cameras are distributed densely enough and with long-term recording capability (e.g., videos or images stored with time and location stamp) is reasonable. If such scenario is not yet a reality, then the abovementioned capability will likely be provided in the near future, as the concept of collaborative sensing \cite{2018Cyber, 2019CrowdTracking, 2017CrowdTracker, Seng2016Heuristics} has emerged, thus greatly enhancing urban sensing ability via the combination of stationary infrastructures and mobile phones. To search within these camera records, particularly for traffic surveillance cameras, the automatic detection of license plates can be utilized; however, for mobile phones, this detection technique may meet difficulties because conditions and qualities are neither uniform nor compliant. In that case, the citizens or policemen themselves determine whether a target object is in a video or an image.

We agree that tracking a person can be more challenging than tracking a car. However, the basic idea behind the proposed technique remains the same: use history trajectories to help in the mobility prediction, then guide the intermediate searching at heuristic moments. To apply our strategy in person tracking, we only need to replace the car trajectories with person trajectories. The difference only lies in the difficulty of single spatiotemporal search, i.e., the unit cost.

\subsection{Future Work}

In the future, we plan to conduct more theoretical analysis of this novel problem, design more efficient strategies (such as based on reinforcement learning), test on more datasets and spatiotemporal granularities, and deduce whether the findings are still applicable if the unit cost of a spatiotemporal search is not a constant but a variable depending on the involved spatiotemporal unit.

An assumption in current work is that we can find the target object from the record definitely if the object was at that spatiotemporal unit. It is a remaining challenge if the record has a possibility to omit the object.

In addition, we will explore a more effective mobility prediction algorithm, as it is also a key factor influencing the total costs of spatiotemporal searches. Particularly, we plan to use fuzzy matching instead of current exact matching to train the prediction model, directly predict the ranking of locations instead of the current probabilities of locations, or use additional information, such as road networks and real-time traffic conditions, to fixate the possible directions and speeds.

% if have a single appendix:
%\appendix[Proof of the Zonklar Equations]
% or
%\appendix  % for no appendix heading
% do not use \section anymore after \appendix, only \section*
% is possibly needed

% use appendices with more than one appendix
% then use \section to start each appendix
% you must declare a \section before using any
% \subsection or using \label (\appendices by itself
% starts a section numbered zero.)
%

%\appendices
%\section{Proof of the First Zonklar Equation}
%Appendix one text goes here.
%
%% you can choose not to have a title for an appendix
%% if you want by leaving the argument blank
%\section{}
%Appendix two text goes here.

% use section* for acknowledgment
\section*{Acknowledgment}

This work is supported by the National Key Research and Development Program of China (No. 2019YFB2102200), the National Science Fund for Distinguished Young Scholars (No. 61725205), the National Natural Science Foundation of China (No. 61772136), and the Fujian Engineering Research Center of Big Data Analysis and Processing.

% Can use something like this to put references on a page
% by themselves when using endfloat and the captionsoff option.
\ifCLASSOPTIONcaptionsoff
  \newpage
\fi

\begin{IEEEbiography}[{\includegraphics[width=1in,height=1.25in,clip,keepaspectratio]{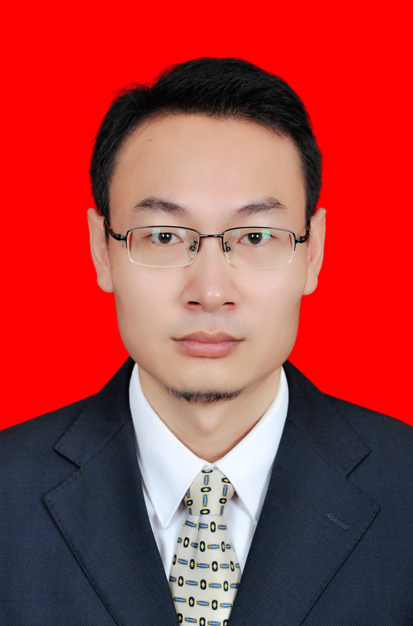}}]{Zhiyong Yu}
	received the M.E. and Ph.D. degrees in computer science and technology from Northwestern Polytechnical University, Xi’an, China in 2007 and 2011, respectively. He is an associate professor in the College of Mathematics and Computer Science, Fuzhou University, Fuzhou, China. He was also a visiting student at Kyoto University, Kyoto, Japan, from 2007 to 2009 and a visiting researcher at the Institut Mines-Telecom, TELECOM SudParis, Evry, France, from 2012 to 2013. His current research interests include pervasive computing, mobile social networks, and mobile crowd sensing.
\end{IEEEbiography}

%\vspace{-100 mm}

\begin{IEEEbiography}[{\includegraphics[width=1in,height=1.25in,clip,keepaspectratio]{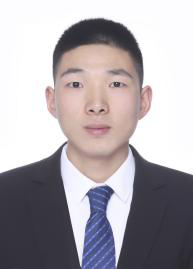}}]{Lei Han}
	received the bachelor's degree and master’s degree in engineering of computer science from Fuzhou University, Fuzhou, China in 2017 and 2020. He is currently a PH.D candidate of computer application technology from Northwestern Polytechnical University, Xi’an, China. His current research interest is pervasive computing and mobile crowdsensing.
\end{IEEEbiography}

%\vspace{-100 mm}

\begin{IEEEbiography}[{\includegraphics[width=1in,height=1.25in,clip,keepaspectratio]{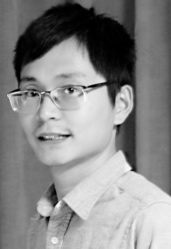}}]{Chao Chen}
	received the B.Sc. and M.Sc. degrees in control science and control engineering from Northwestern Polytechnical University, Xi’an, China in 2007 and 2010, respectively, and the Ph.D. degree in computer science from Pierre and Marie Curie University and Institut Mines-Télécom/Télécom Sud-Paris, France in 2014. He is currently a full professor with the College of Computer Science, Chongqing University, Chongqing, China. He has authored or coauthored more than 80 papers including 20 ACM/IEEE Transactions. His research interests include pervasive computing, mobile computing, urban logistics, data mining from large-scale taxi GPS trajectory data, and big data analytics for smart cities.
\end{IEEEbiography}

%\vspace{-100 mm}

\begin{IEEEbiography}[{\includegraphics[width=1in,height=1.25in,clip,keepaspectratio]{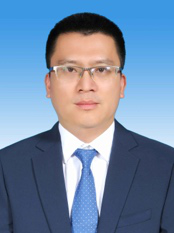}}]{Wenzhong Guo}
	received his Ph.D. degree in the Department of Physics and Information Engineering, Fuzhou University, Fuzhou, China in 2010. He is a professor and director of Fujian Key Laboratory of Network Computing and Intelligent Information Processing. He was a postdoctoral researcher in the Department of Computer Science, National University of Defense and Technology, Changsha, China from 2011 to 2014, a visiting professor at the Faculty of Engineering, Information and System, University of Tsukuba, Japan in 2013, and a visiting professor at the Department of Computer Science and Engineering, State University of New York at Buffalo, USA in 2016. His research interests include wireless sensor networks, network computing, and mobile crowdsensing.
\end{IEEEbiography}

%\vspace{-140 mm}

\begin{IEEEbiography}[{\includegraphics[width=1in,height=1.25in,clip,keepaspectratio]{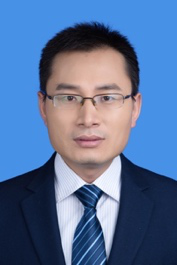}}]{Zhiwen Yu}
	received the M.E. and Ph.D. degrees from Northwestern Polytechnical University, Xi’an, China in 2003 and 2005, respectively. He is a professor in the School of Computer Science, Northwestern Polytechnical University. He visited the Institute of Information and Communication in Singapore from 2004 to 2005. From 2006 to 2009, he was a postdoctoral researcher at Nagoya University and a special researcher at Kyoto University  in Japan. From November 2009 to October 2010, he was funded by the German Humboldt Foundation and went to the University of Mannheim in Germany for collaborative research. His current research interests include pervasive computing, mobile crowdsensing, internet of things, and intelligent information technology.
\end{IEEEbiography}

% that's all folks
\end{document}